# Automating Attendance Management in Human Resources: A Design Science Approach Using Computer Vision and Facial Recognition

Bao-Thien Nguyen Tat[a,b], Minh-Quoc Bui[a,b], Vuong M. Ngo[c,*]

[a]*University of Information Technology, Ho Chi Minh City, Vietnam*
[b]*Vietnam National University, Ho Chi Minh City, Vietnam*
[c]*Ho Chi Minh City Open University, Ho Chi Minh City, Vietnam*

**Abstract**

Haar Cascade is a cost-effective and user-friendly machine learning-based algorithm for detecting objects in images and videos. Unlike Deep Learning algorithms, which typically require significant resources and expensive computing costs, it uses simple image processing techniques like edge detection and Haar features that are easy to comprehend and implement. By combining Haar Cascade with OpenCV2 on an embedded computer like the NVIDIA Jetson Nano, this system can accurately detect and match faces in a database for attendance tracking. This system aims to achieve several specific objectives that set it apart from existing solutions. It leverages Haar Cascade, enriched with carefully selected Haar features, such as Haar-like wavelets, and employs advanced edge detection techniques. These techniques enable precise face detection and matching in both images and videos, contributing to high accuracy and robust performance. By doing so, it minimizes manual intervention and reduces errors, thereby strengthening accountability. Additionally, the integration of OpenCV2 and the NVIDIA Jetson Nano optimizes processing efficiency, making it suitable for resource-constrained environments. This system caters to a diverse range of educational institutions, including schools, colleges, vocational training centers, and various workplace settings such as small businesses, offices, and factories. Its adaptability to distinct organizational requirements ensures its relevance and effectiveness across a broad spectrum of users. One of the distinguishing features of this system is its robust integration with databases. It enables efficient storage of attendance records and supports customizable report generation. This comprehensive data management capability ensures that attendance data is readily accessible for monitoring and analysis purposes, contributing to improved decision-making processes. Implementing this Haar Cascade-based attendance management system offers several significant benefits. It not only reduces the manual workload associated with attendance tracking but also minimizes errors, enhancing overall accuracy. The system's affordability and efficiency democratize attendance management technology, making it accessible to a broader audience. Consequently, it has the potential to transform attendance tracking and management practices, ultimately leading to heightened productivity and accountability. In conclusion, this system represents a groundbreaking approach to attendance tracking and management. By combining Haar Cascade, OpenCV2, and the NVIDIA Jetson Nano, it addresses the specific needs of educational institutions and workplaces, offering a cost-effective, efficient, and adaptable solution that has the potential to revolutionize attendance management practices.

*Correspondence: Vuong M. Ngo, email: vuong.nm@ou.edu.vn

## 1. Introduction

In a world where technological innovations continue to redefine the boundaries of management and information systems, our research serves as a pivotal bridge between the theoretical realms of Management Information Systems (MIS) and practical implementation in challenging contexts. By introducing an innovative attendance management system founded on the Haar Cascade technique, we not only address a pervasive gap in the literature but also offer a tangible solution to real-world problems faced by organizations and institutions in low-resource settings.

A Face Recognition Attendance Management System (FRAMS) is an innovative approach to attendance tracking that utilizes the latest facial recognition technology to identify and record the presence of individuals. This technology can identify individuals in a fraction of a second, making the attendance process fast and efficient. The system uses a camera to capture an image of the face, and then compares it to a database of pre-registered images to determine the identity of the individual. This technology is not only faster than traditional attendance methods, but it also eliminates the need for manual record-keeping and reduces the likelihood of errors. With its high level of accuracy and ease of use, the FRAMS is becoming an increasingly popular solution for attendance tracking in a wide range of settings, from educational institutions and workplaces to events and public spaces.

However, this innovation didn't emerge in a vacuum; it addresses persistent challenges and limitations inherent in traditional attendance methods. To offer a clear context for the reader, let's explicitly delve into the problem statement, elucidating the shortcomings and challenges that our FRAMS system aims to resolve.

Our research paper represents a pioneering step forward in the realm of Management Information Systems. It offers a practical and innovative approach to attendance management that holds relevance in a wide range of scenarios. Unlike conventional attendance systems, which are often costly and impractical in resource-constrained environments, our work focuses on exploring the technical intricacies, performance benchmarks, and scalability aspects of our system. By addressing these critical factors, we provide a novel and highly efficient solution to attendance management, bridging the gap between theory and practical application.

At its core, our research unveils an innovative and highly efficient solution for attendance management, redefining established paradigms within MIS. While attendance systems are not a novel concept in this field, our approach distinctively marries cutting-edge technology, affordability, and scalability, addressing critical challenges faced by organizations, particularly in education and corporate environments, thereby reshaping the landscape of MIS practices.

A pivotal aspect of our research lies in its commitment to accessibility in resource-constrained settings, which is of paramount importance in MIS. Many educational institutions and smaller organizations encounter hurdles in accessing high-end hardware or expensive facial recognition solutions. To alleviate this, our system has been meticulously engineered to operate seamlessly on cost-effective hardware, such as the Raspberry Pi and NVIDIA Jetson Nano. This strategic emphasis on accessibility not only broadens the scope of advanced technology but also underscores its pivotal role within MIS as an inclusive and transformative solution.

Furthermore, our research meticulously addresses privacy concerns, a critical facet within MIS. We have rigorously ensured that our system captures and stores only essential information required for attendance purposes, while implementing stringent data protection measures that align with evolving regulatory frameworks, setting a benchmark for responsible and ethical data management practices within MIS.

Additionally, the scalability and practical implementation of our system are pivotal elements that resonate deeply with MIS practices. Our solution efficiently handles attendance management across a spectrum of environments, catering to both small and large groups. Our research extends beyond theoretical concepts to provide actionable insights into implementing such systems, offering guidance on data collection, training, real-world testing, and seamless system integration, thereby enriching MIS practices with practical knowledge invaluable for organizations contemplating the adoption of similar solutions.

Our research spans the boundaries of MIS, reflecting its interdisciplinary relevance. By seamlessly integrating hardware, software, and image processing techniques, it extends its influence not only within MIS but also into broader domains such as computer science, artificial intelligence, and engineering. This interdisciplinary versatility underscores the profound significance of our solution in MIS, reinforcing its role as a transformative and versatile asset with a wide range of applications, enriching MIS practices and contributing to technological advancements across various sectors.

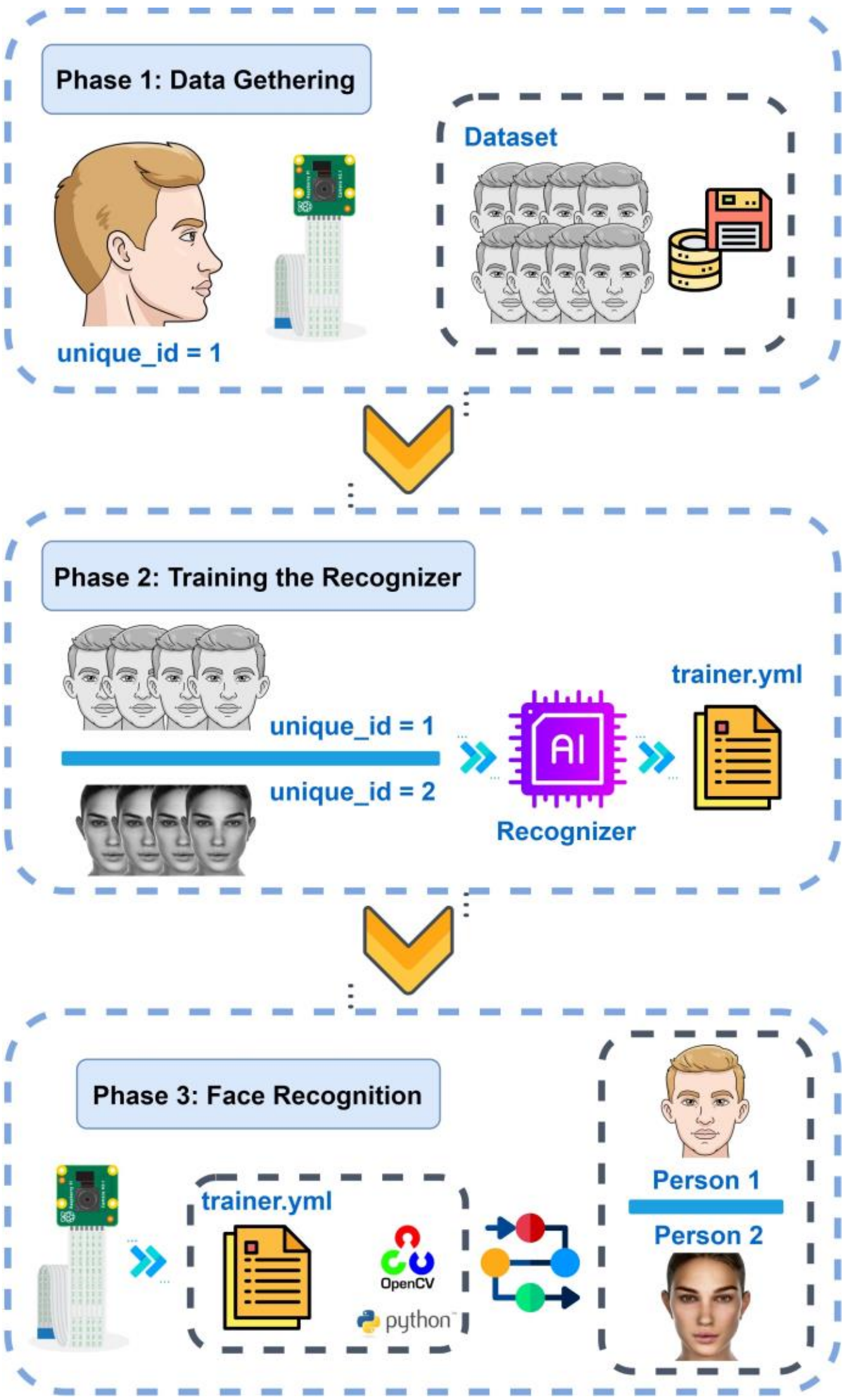

Fig. 1. Diagram of the main operation of the recognition algorithm.

The landscape of attendance tracking has long grappled with multifaceted challenges stemming from conventional methods, including verbal roll call, fingerprint scanning, smart cards, and self-registration systems. Each of these methods presents its own unique set of advantages and disadvantages, catering to the varying needs of different organizations:

- Traditional attendance-taking, characterized by the verbal call of students' names, is not only a time-consuming endeavor but also prone to human error, leaving ample room for discrepancies.
- Fingerprint scanning, while lauded for its security measures and resistance to forgery, imposes its own set of limitations. It restricts individuals from attempting proxy attendance on behalf of others or engaging in fraudulent activities. However, the deployment of fingerprint scanning systems often necessitates substantial investment and can be unwieldy in certain settings.

- Smart card-based attendance solutions find favor among large organizations, where the management of numerous employees necessitates streamlined processes. Nevertheless, this method introduces its own set of complexities, as employees may forget their cards, misplace them, or inadvertently clock in for their colleagues. Additionally, the acquisition costs for cards and associated machinery can be prohibitive.

To bridge the narrative seamlessly between this contextualization and the specific focus of our study, it is prudent to include a transitional paragraph. This paragraph would serve to clearly communicate our study's objective, which is to address the limitations inherent in existing attendance methods through the implementation of FRAMS, a cutting-edge solution driven by artificial intelligence and powered by the NVIDIA Jetson Nano embedded computer.

Our research paper's primary objective is to present the various facets of our attendance management system, which leverages the formidable capabilities of Haar Cascade and OpenCV2. These technologies enable our system to efficiently process video streams, exhibit robustness in the face of variations in lighting and pose, all while maintaining minimal computational requirements. Furthermore, our study seeks to underscore the potential of embedded computers like the NVIDIA Jetson Nano in delivering energy-efficient, cost-effective solutions for attendance tracking. The synergy between hardware and software components serves as a cornerstone in achieving our system's seamless integration.

One of the core strengths of our work lies in its unique contribution to the field. While we briefly allude to the use of NVIDIA Jetson Nano and artificial intelligence technology, we will now explicitly outline how our research uniquely addresses a gap in the existing literature. This gap is characterized by the absence of comprehensive solutions that blend cutting-edge technology, energy efficiency, and cost-effectiveness for attendance tracking. Our system's innovative approach bridges this gap, positioning it as a pioneering solution in the field of attendance management.

Furthermore, our research is particularly unique in its approach, as it places a strong emphasis on drawing inspiration and foundational knowledge from well-established, reputable articles. This process of referencing and building upon credible sources plays a pivotal role in shaping the framework and context of our study. Expanding further upon this approach, our commitment to anchoring our research in established literature is paramount in ensuring the robustness of our methodology and the credibility of our findings. By delving deep into the extensive body of work represented by references (Alter et al., 2022; Ashok et al., 2022; Koohang et al., 2022; Namvar et al., 2023; Samuel et al., 2022; Tran et al., 2023; Johnson et al., 2022; Zeuge et al., 2023; Pan et al., 2023; Costa-Climent R et al., 2023; Islam et al., 2023; Iqbal et al., 2019; Kavitha et al., 2023; Kim et al., 2021; Liu et al., 2021; Muyambo et al., 2018; Okokpujie et al., 2018; Raju et al., 2021; Rusia et al., 2019; Santhoshkumar et al., 2019; Selvi et al., 2019; Sharma et al., 2022; Yawar et al., 2022; Zhang et al., 2021), we not only draw upon the wisdom and insights of seasoned experts but also contribute to the ongoing conversation within our field. This iterative process of referencing, reviewing, and synthesizing the wealth of knowledge encapsulated in these reputable sources serves as the bedrock upon which our research stands. Moreover, it not only affords our study a sense of academic rigor but also fosters a rich and interconnected tapestry of ideas that enriches our own work.

In essence, our research stands out because of its dedication to leveraging the wealth of knowledge found in reputable articles, ultimately allowing us to make a meaningful contribution to our chosen field by building upon the strong foundations laid by those who came before us: we built and developed an attendance management system using face recognition technology with NVIDIA Jetson Nano embedded computer aims to replace the manual attendance and fingerprint/card-based timekeeping methods, which still have many drawbacks, with a fully automatic system based on artificial intelligence technology. Fig. 1 shows the prototype of our attendance management system, which includes a camera for capturing facial images and the NVIDIA Jetson Nano for processing the images and recognizing the faces.

Our paper presents four contributions, which are as follows:

- Developing a reliable and efficient attendance management system using state-of-the-art technologies, which can be deployed in various settings for attendance tracking and monitoring.
- Demonstrating the effectiveness of using Haar Cascade and OpenCV2 for building an attendance management system, which offers efficiency in processing video streams, robustness to variations in lighting and pose, and low computational requirements.
- Showing the potential of an embedded computer like NVIDIA Jetson Nano in providing an energy-efficient and cost-effective solution for attendance tracking, with the ability to integrate hardware and software components.
- Contributing to the advancement of facial recognition technology and its practical applications in attendance management systems, paving the way for wider adoption in various settings.

The paper is structured to provide a comprehensive and systematic exploration of our research topic. Initially, in Section 2, we present an extensive review of related works, exploring the contributions and findings of other researchers in the field. This sets the stage for understanding the current landscape and identifying gaps that our study aims to address. Following this, Section 3 delves into the theoretical frameworks underpinning our research. This section discusses the background theories and methodologies, including the Design Science Research Methodology and Paradigmatic Analysis of Information Systems as a Design Science, which guide our approach and lend rigor to our study. In Section 4, we detail our research methodology, outlining the steps and processes employed in developing and implementing our face recognition attendance management system. This includes an in-depth look at the structural and software design, as well as the overall system development process. Section 5 is dedicated to our experimental design and evaluation. Here, we present the implementation of our research model, the methodology for data collection and preparation, and a thorough analysis of the system's performance. This section critically evaluates the effectiveness and efficiency of our system, providing a clear understanding of its capabilities and limitations. The paper then progresses to Section 6, where we discuss the broader implications of our findings. This section offers an in-depth analysis of the results, exploring their significance in the context of the existing literature and theoretical frameworks. Finally, the paper concludes in Section 7, where we summarize the key contributions of our study and propose directions for future research. This closing section also acknowledges the valuable contributions that have informed and supported our research journey.

## 2. Literature Review

In recent years, the significance of face recognition systems has escalated in various applications such as surveillance, security, and biometrics. With the advancement of deep learning techniques, many researchers have concentrated on developing neural network-based face recognition systems. However, the high computational demands of these systems often make them impractical for embedded systems. In this context, the groundbreaking approach of Y. Wen et al. (2021) by using a deep convolutional neural network (DCNN) for face recognition marks a significant advancement. Their method's ability to efficiently extract discriminative facial features while achieving state-of-the-art performance on benchmark datasets like LFW, YTF, and IJB-A, and at a reduced computational cost, emphasizes the feasibility of efficient face recognition and extends its applicability.

The realm of embedded systems has experienced a surge in real-time object detection and tracking applications. Innovations like those by Mani et al. (2021), leveraging the Raspberry Pi Camera and OpenCV, utilize the YOLOv4 neural network architecture, demonstrating impressive accuracy on a cost-effective, low-power hardware platform. Similarly, the system of Singh et al. (2021) for vehicle detection using the Haar Cascade Classifier highlights the computational simplicity and high accuracy achievable in embedded systems for real-time applications.

The importance of hardware selection in face recognition systems is underscored by comparative studies such as those by Salih et al. (2020). Their work reveals a significant performance differential between the NVIDIA Jetson Nano and Raspberry Pi in real-time recognition tasks, indicating the crucial role of appropriate hardware in optimizing system performance.

In 2021, contrasting performances were observed in studies utilizing the Haar Cascade algorithm for face recognition. Adoghe et al. (2021) implemented this algorithm on the Raspberry Pi, supported by the Dlib library and OpenCV, but encountered limitations with an accuracy rate of only 72.9%. This highlights the nuances of hardware selection and its impact on system performance. Conversely, Chandramouli (2021) developed an attendance management system using the Haar Cascade algorithm on the NVIDIA Jetson Nano, benefiting from the additional GPU support provided by NVIDIA's TensorRT. These studies underline the importance of selecting the right hardware platform for specific tasks in embedded systems.

The embedded face recognition domain continues to face the challenge of balancing accuracy with computational efficiency. Nayak et al. (2021) development of a real-time face recognition system using Haar Cascade classifiers on the NVIDIA Jetson Nano is a testament to this balance, achieving high accuracy with minimal resource consumption. This study not only illustrates the advancements in embedded systems but also sets the stage for exploring how AI and ML are optimizing various facets of modern life.

Moving from the technical intricacies of embedded systems to the human-centric application in workplaces, the literature reveals a significant shift towards AI-enhanced processes. Mittal, Mahendra, Sanap and Churi (2022) delve into the realm of stress management, applying machine learning to identify stress factors in educational and professional settings. Their research exemplifies the potential of AI in improving mental health and workplace productivity. Complementing this, Votto, Valecha, Najafirad and Rao (2021) shed light on the integration of AI in Tactical Human Resource Management, marking a transformation in HR practices towards more efficient, data-driven operations.

Security, a paramount concern in the modern world, is also being reshaped by AI, as highlighted by Khan and Efthymiou (2021). Their assessment of biometric technology at airports, particularly within the U.S. Customs and Border Control's Biometric Entry Exit Program, emphasizes AI's expanding role in enhancing national security, showcasing its application in verifying identities and detecting document fraud.

In the context of urban development, AI's influence is further broadened, as illustrated by Herath and Mittal (2022). Their comprehensive review of AI adoption in smart cities explores its impact across sectors, from healthcare to transportation, portraying AI as a key driver in making cities more efficient and livable. This narrative extends to the cybersecurity domain, where Lata and Singh (2022) highlight AI's critical role in protecting digital infrastructures through advanced intrusion detection systems.

Moreover, the versatility of AI is underscored in the field of human activity recognition, with Ray et al. (2023) discussing the significance of transfer learning in vision-based systems. This research not only demonstrates AI's adaptability but also its ability to enhance accuracy and reduce data scarcity challenges.

In the manufacturing industry, the transformative power of AI is evident in the work of Jamwal et al. (2022), who explore deep learning applications within Industry 4.0, highlighting AI's role in promoting sustainable manufacturing practices. Additionally, Pathare et al. (2023) challenge conventional data generation methods in AI, advocating for innovative approaches in synthetic data generation.

Finally, the narrative culminates in addressing one of the most pressing issues in AI: bias management. Dr. Varsha (2023) provides a systematic review of biases in AI systems, emphasizing the need for ethical and responsible AI practices. This study not only highlights the challenges in AI development but also underscores the importance of maintaining fairness and integrity in technology deployment.

In essence, this collection of literature weaves a narrative that highlights both the challenges and breakthroughs in AI and ML. It portrays a journey from optimizing computational efficiency in face recognition systems to enhancing workplace productivity, securing national borders, and advancing urban living, all while navigating the ethical complexities inherent in AI development.

As we navigate the landscape of embedded face recognition, it becomes increasingly apparent that staying at the forefront of technology trends and addressing associated challenges is paramount. The intersection of embedded systems, AI, and ML continues to offer innovative possibilities, shaping the future of face recognition and its multifaceted applications in various domains. Numerous methods and devices have emerged for the development of face recognition attendance management systems, but a central question remains: How can we construct a face recognition system tailored for mobile devices characterized by small form factors, low power consumption, affordability, real-time processing, and robust safety and security measures?

The preceding works collectively illuminate the path toward realizing such a vision. By harnessing the Haar Cascade classifiers in conjunction with the NVIDIA Jetson Nano, these studies demonstrate the practicality of deploying face recognition systems under stringent resource constraints. Achieving high accuracy and real-time performance within these resource limitations is a testament to their efficacy. Importantly, these works offer invaluable insights into the intricate facets of system design, implementation strategies, and meticulous performance evaluations, forming a foundational knowledge base for the development and enhancement of face recognition systems optimized for mobile and embedded platforms.

Towards the end of this literature review, it is imperative to provide a brief justification for the approach proposed in our research. We have chosen to combine the processes of Detection, Extraction, Labeling, and Face Recognition using the Haar Cascade algorithm and NVIDIA Jetson Nano for several compelling reasons. Firstly, our approach aligns with the findings from multiple studies reviewed in this section. The Haar Cascade algorithm offers a computationally efficient method for face detection, a critical component in face recognition systems. By utilizing the NVIDIA Jetson Nano, we capitalize on the capabilities of specialized AI embedded hardware, known for its real-time processing performance and suitability for resource-constrained environments.

Secondly, our approach builds on the reviewed studies by addressing a specific need – the development of a comprehensive, yet resource-efficient, face recognition system tailored for mobile and embedded systems. While the individual studies emphasize various aspects of face recognition, our approach unifies these aspects into a cohesive system that encompasses detection, feature extraction, labeling, and recognition. This holistic integration streamlines the entire process, optimizing for both accuracy and efficiency.

In summary, our approach synthesizes the insights gained from these related works into a unified framework. It not only leverages the strengths of the Haar Cascade algorithm and the NVIDIA Jetson Nano but also advances the field by proposing a comprehensive solution tailored to the unique demands of mobile and embedded systems. Based on the ideas and experimental results of the authors (Salih et al., 2020; Adoghe et al., 2021; Chandramouli, 2021; Nayak et al., 2021), we propose combining the process of Detection, Extraction, Labeling, and Face Recognition using the Haar Cascade algorithm with moderate complexity and high accuracy on the NVIDIA Jetson Nano embedded computer. This approach takes advantage of specialized AI embedded computers' ability to provide high real-time processing performance, running a large volume of AI processing workloads through 472 GFLOPs capable of quickly running modern deep learning algorithms.

## 3. Theoretical Frameworks

In the landscape of designing and implementing an efficient face recognition attendance system, the theoretical underpinnings play a crucial role in guiding the research methodology and ensuring the practical relevance of the system. This section delves into the theoretical frameworks that form the backbone of our study, including the Design Science Research Methodology, Paradigmatic Analysis of Information Systems, and the technical foundation of Haar Cascades and their detection in OpenCV.

### *3.1. Design Science Research Methodology*

Our research meticulously follows the Design Science Research Methodology (DSRM) as proposed by Peffers et al. (2007), which is pivotal in the realm of information systems research. DSRM is a methodological blueprint for creating and evaluating IT artifacts that aim to solve identified problems. In the context of our study, this framework serves as a structured guide through various research stages, from the inception of the idea to its practical realization.

DSRM's iterative process begins with the identification and motivation of a problem. In our case, this involved recognizing the need for an efficient, accurate, and cost-effective face recognition system for attendance management. The next phase, defining the objectives of a solution, shaped our pursuit of a system that integrates embedded computing with advanced face recognition technologies.

The design and development stage, which is the core of DSRM, involves creating the actual artifact in our context, the face recognition attendance system. This stage demanded a deep dive into both hardware and software components, ensuring that each part synergistically contributes to the overall functionality and effectiveness of the system.

Supporting our approach, Barata et al. (2023) discuss the strategic importance of aligning research objectives with project stakeholders' expectations, an aspect we meticulously addressed by engaging with key stakeholders from the educational and organizational domains to refine our system's objectives and functionalities.

The demonstration phase in DSRM, where the artifact is tested in a real-world scenario, is critical for validating the utility of our system. For our research, this involved deploying the system in an educational or organizational setting to track attendance accurately and efficiently. Humble and Mozelius (2023) emphasize the adaptability of DSRM for small-scale and context-specific studies, which was instrumental in our pilot testing phase where initial user feedback was gathered and analyzed to gauge the system's performance and usability.

The evaluation stage, arguably one of the most crucial, involves assessing the artifact against predefined criteria. Our system's evaluation was multifaceted, examining its accuracy, user-friendliness, and efficiency. This stage is instrumental in refining the system, ensuring it meets the desired standards. Weigand and Johannesson (2023) provide guidelines for specifying and identifying DSR artifacts in IS research, which helped in precisely defining the functional and non-functional requirements of our face recognition system.

Finally, DSRM emphasizes the importance of communication, advocating for the dissemination of findings within the scientific community. Our research aligns with this principle, aiming to contribute valuable insights to the field of face recognition systems in attendance management. Through the integration of recent advancements and critiques of DSRM, we aim not only to apply this methodology but also to contribute to its ongoing discourse and development in the academic community.

### *3.2. Paradigmatic Analysis of Information Systems as a Design Science*

Complementing the methodological rigor of DSRM, Iivari's (2007) Paradigmatic Analysis of Information Systems as a Design Science offers a theoretical perspective that enriches our research. This analysis underscores the significance of developing practical and innovative solutions within the framework of information systems. It provides a lens through which the creation, implementation, and impact of IT artifacts are examined, emphasizing their contribution to both theory and practice.

In applying Iivari's paradigmatic analysis, our research is positioned at the intersection of theoretical innovation and practical application. The development of our face recognition attendance system is not just a technical achievement but also a contribution to the body of knowledge in information systems. This system, while solving a practical problem, also embodies the principles of design science, which values creativity, utility, and innovation.

Our research, through the application of these paradigms, addresses the broader objectives of information systems: the creation of systems that are not only technologically advanced but also socially relevant and contextually appropriate. By focusing on an attendance system that leverages face recognition technology, we are responding to real-world needs while also pushing the boundaries of what is technologically feasible in the domain of embedded systems and AI.

### *3.3. The Theory of Haar Cascades*

In 2001, Viola and Jones (2001) introduced the Haar feature-based cascade classifiers as a novel object detection method that could be applied to various tasks, including face detection. The proposed method used a cascade of simple features to efficiently and accurately detect objects in images, with a focus on real-time performance. The paper presented the results of experiments that demonstrated the effectiveness of the method in detecting faces, as well as other objects such as pedestrians and vehicles, with high accuracy and speed. This paper has been cited extensively in the computer vision literature and has had a significant im-pact on the development of object detection and recognition systems. This ma-chine learning approach involves training a cascade function on numerous positive and negative images to detect objects in other images. Haar feature-based cascade classifiers have proven to be an effective technique for object detection.

In this particular case, our focus is on detecting faces. To begin with, they re-quire a large number of positive images (images containing faces) and negative images (images not containing faces) to train the classifier. Once that is done, they need to extract features from these images. They use Haar features, which are similar to convolutional kernels, to accomplish this. Each feature is represented by a single value obtained by subtracting the sum of pixels under the white rectangle from the sum of pixels under the black rectangle. Fig. 2 shows different types of Haar-like features that can be extracted from an image patch using the Haar Cascade algorithm, including edge features, line features, and corner features. These features are used to detect different types of objects in an image, and the Haar Cascade algorithm combines them into a strong classifier that can detect object with high accuracy.

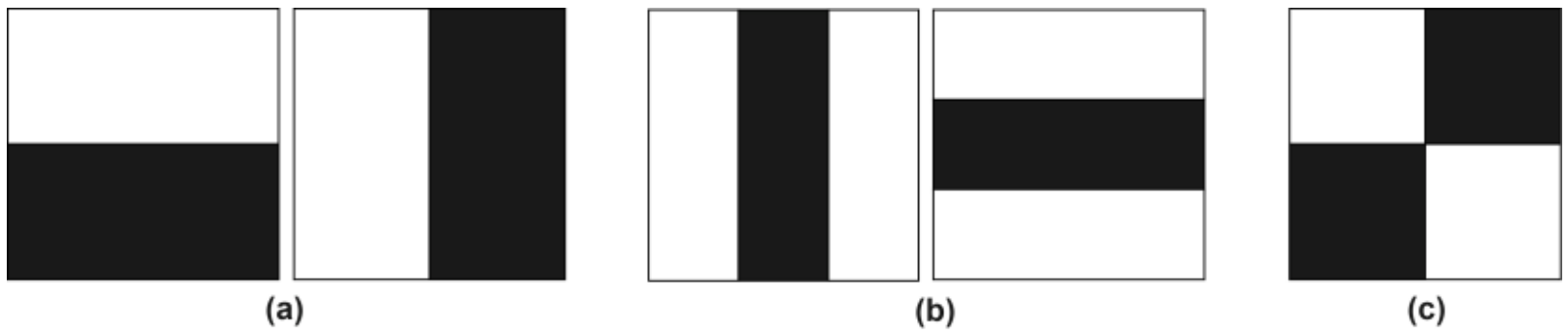

Fig. 2. Different types of Haar-like features extracted from an image patch, (a) Edge Features, (b) Line Features, (c) Four-rectangle Features.

The algorithm involves computing a multitude of features by utilizing all possible sizes and positions of each kernel. This requires a significant amount of computation power, especially for a 24x24 window that yields over 160,000 features. To calculate each feature, the sum of the pixels under the white and black rectangles needs to be determined. To simplify this process, the integral image was introduced. This technique reduces the number of calculations for a specific pixel to just four pixels, regardless of the image's size. This makes the computation significantly faster and more efficient

Out of all the features that are calculated, a significant number of them are not useful. To illustrate, take a look at the image below. The first and second features on the top row are good examples of useful features. The first feature focuses on the observation that the region around the eyes is usually darker than the nose and cheeks. The second feature relies on the notion that the eyes are darker than the bridge of the nose. However, applying the same windows to the cheeks or other areas is not relevant. Therefore, how do they pick the most features out of the over 160,000 features? AdaBoost is used to accomplish this task.

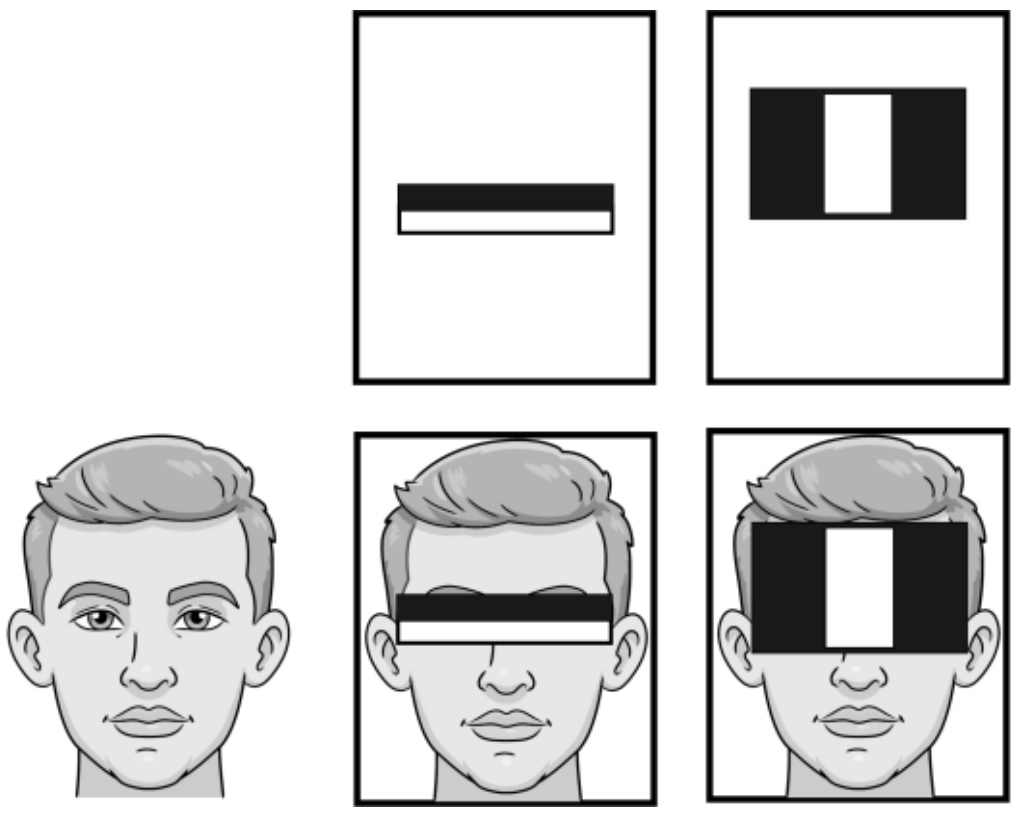

Fig. 3. Haar Cascades are a class of object detection algorithms.

To select the most effective features for classification, each feature is applied to all the training images, and a threshold is determined for each feature that best distinguishes between positive and negative examples. However, misclassifications will inevitably occur, and the features with the lowest error rates are selected as the most informative. This selection process is not straightforward. Initially, each image is assigned an equal weight, but after each classification round, the weights of misclassified images are increased. This is repeated until the desired level of accuracy or the required number of features has been obtained. Fig. 3 shows an example of the algorithm detecting faces in an image. The rectangles represent the detected faces.

The weak classifiers are combined to create a final classifier through a weighted sum. Though the individual weak classifier alone is not capable of classifying the image, the final combined result forms a strong classifier. According to the paper, event 200 features can yield a 95% accuracy in detection. The final setup consisted of a round 6,000 features, representing a substantial reduction from the original 160,000+ features.

$$h(x) = sgn\left(\sum_{j=1}^{M} \alpha_j \, h_j(x)\right) \tag{1}$$

Each weak classifier is a threshold function based on the feature $f_j$.

$$h_j(x) = \begin{cases} -s_{j,}\ if\ \ f_j < \theta_j \\ s_{j,}\ otherwise \end{cases} \tag{2}$$

To optimize face detection, it is wise to have a quick and easy way to determine if a window does not contain a face in an image. If it is confirmed that a region is not a face, then it can be immediately rejected and not processed again. Instead, the algorithm can concentrate its efforts on areas that have a higher probability of containing a face. By using this approach, the majority of the image, which is typically non-face regions, can be disregarded, thereby saving computational resources and increasing efficiency.

They devised the Cascade of Classifiers technique to address this issue. Rather than employing all 6,000 features on a window, the features are split into various stages of classifiers and run sequentially (usually, the initial few stages will have considerably fewer features). If a window fails the first stage, it is immediately discarded, and the remaining features are not applied. However, if it passes, the next stage of features is applied, and the process continues. A window that passes all stages is classified as a face region.

The detector developed by the authors consisted of 38 stages, each utilizing a varying number of features, totaling over 6,000. The first five stages contained 1, 10, 25, 25, and 50 features, respectively, and the two features shown in the previous image were selected as the top performers through AdaBoost. On average, only 10 features out of the 6000+ were evaluated for each sub-window, according to the authors.

### *3.4. Haar-cascade Detection in OpenCV*

Haar-Cascade detection is a popular object detection used in computer vision applications. OpenCV, an open-source computer vision library, provides an implementation of this technique that can be used to detect objects such as faces, eyes, and pedestrians.

The detection process involves training a classifier with positive and negative images of the object. The positive images are samples of the object, while the negative images do not contain the object. The classifier is then used to detect the object in new images.

OpenCV provides a pre-trained classifier for detecting faces, which can be easily used in Python code. The process involves loading the classifier file, capturing images from a camera or video, and passing the image frames to the classifier. If a face is detected, the classifier returns the coordinates of the face in the images.

Haar-Cascade detection is a lightweight technique compared to deep learning-based object detection techniques, making it suitable for low-power devices like Raspberry Pi and NVIDIA Jetson Nano. Its simple implementation and low computational requirements also make it easier to program and less expensive than deep learning techniques.

Despite the availability of more advanced deep learning algorithms, Haar Cas-cade classifiers are still a viable option for object detection and recognition on the NVIDIA Jetson Nano. One reason for this is the relatively lower processing power required by Haar Cascades compared to deep learning algorithms, which can be advantageous for resource-constrained devices like the NVIDIA Jetson Nano. Additionally, Haar Cascades can be quicker to train compared to deep learning models, which can be useful in scenarios where training time is a constraint. Moreover,

Haar Cascades can be more interpretable compared to deep learning models, making it easier to understand how the algorithm is making decisions. This can be important for applications that require explainability and transparency. Finally, Haar Cascades can be more robust to changes in environment or lighting conditions compared to deep learning models, which can be sensitive to variations in the data. While deep learning algorithms are still the go to choice for many computer vision tasks, Haar Cascades remain a useful and viable option for certain applications on the NVIDIA Jetson Nano.

In the OpenCV library, there is a collection of pre-trained Haar Cascades available for use. These cascades are primarily used for detecting different parts of the body such as the face, eyes, mouth and full or partial body.

One option for detecting faces in images is to use OpenCV, which offers a training method (refer to Cascade Classifier Training) as well as pre-trained models that can be loaded using the `cv::CascadeClassifier::load` method. These models are included in the data folder of the OpenCV installation.

## 4. Methodology

This research study is based on key concepts that have been thoroughly explained, and related works have been extensively analyzed to provide a basic understanding of the study's purpose. In this section, detailed information is provided on the system design and methodology used to model the research study's objectives, as well as practical considerations for implementation in real-life scenarios.

### *4.1. The Structural Design*

The NVIDIA Jetson Nano, a cost-effective AI computer, was utilized in this study. This device offers high performance for modern AI workloads, despite its small size, and consumes only 5 watts of power, making it power-efficient. The Jetson Nano has dimensions of 80mm x 100mm and is equipped with a 128-core Maxwell GPU, quad-core Arm Cortex-A57 CPU, and 4GB system memory. Fig. 4 shows the NVIDIA Jetson Nano board with its various components clearly visible, including GPU, CPU and memory. The board also includes various port for connecting external devices such as cameras, dis-plays, and other peripherals that make it become a versatile tool for AI applications.

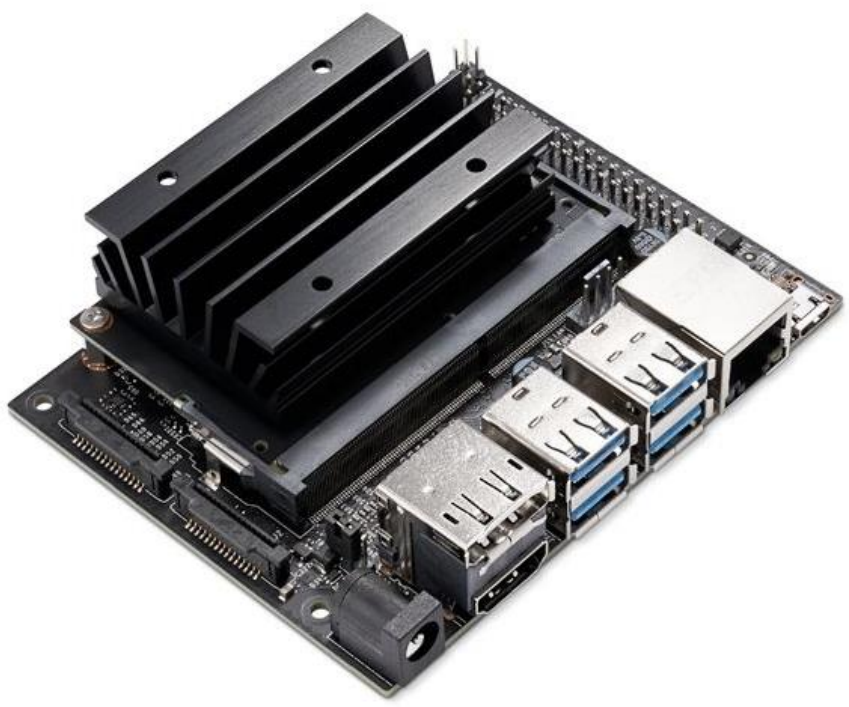

Fig. 4. NVIDIA Jetson Nano.

The Raspberry Pi Camera Module V2 is a versatile camera that has been designed to work specifically with embedded computer. With its Sony IMX219 8-megapixel sensor, it can capture images with great detail and clarity. Additionally, the camera module is capable of recording high-quality videos at a resolution of 1080p30 or 720p60. Fig. 5 shows the camera module with its attached ribbon cable, ready to be connected to an embedded computer for use in an application such as our attendance management system using face recognition technology.

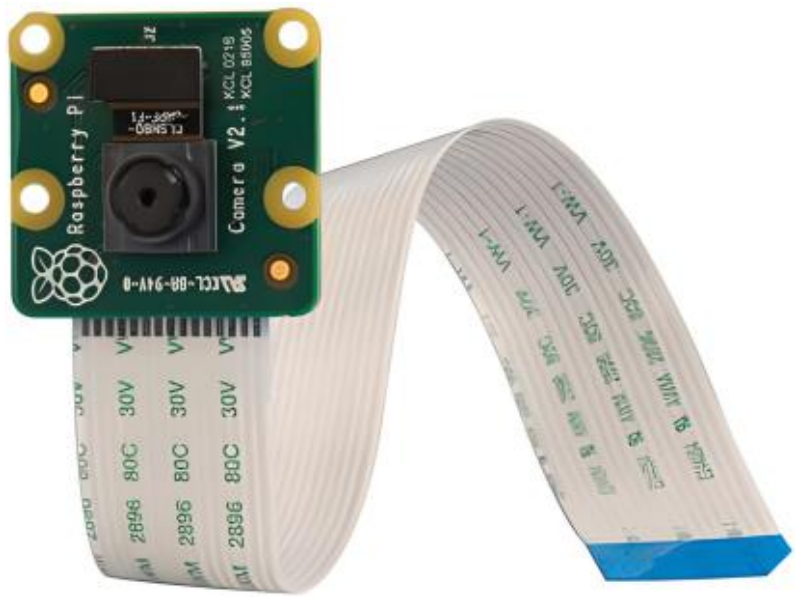

Fig. 5. Raspberry Pi Camera Module V2.

### *4.2. Software Design Specification*

Python, a popular high-level programming language, is commonly used for soft-ware development in various domains, such as data science, web development, and machine learning. Python has powerful libraries that enable developers to build efficient and effective software applications. Visual Studio Code is a widely used integrated Development Environment (IDE) that provides support for multiple programming languages, including Python. It offers many advanced features that can aid in software development, such as debugging, syntax highlighting, and code completion. Together, Python and Visual Studio Code provide robust platforms for developing high-quality software applications with detailed software design specifications.

OpenCV2, a Python library for real-time computer vision, is an important tool for software developers in the field of computer vision. With OpenCV2, developers can access wide range of computer vision functions, such as image and video processing, feature detection, and object recognition. When combined with Python's powerful functional libraries, developers can create high-performance software systems for real-world applications. By using Visual Studio Code and Python with OpenCV2, developers can create powerful software systems with ease.

### *4.3. System Development Process*

To enable the tracking and recognition of face images, the Pan-Tilt Raspberry Pi Camera prototype was developed using the following steps:

- The person need to be identified will be captured through the Raspberry Pi Camera by standing and looking straight into it.
- A folder containing images of the person need to be recognized, collected and labelled beforehand according to the syntax: `User.[Unique Index].[Sequence Index].`
- When the recognition process takes place, the captured frame from the real-time through Raspberry Pi Camera of the person need to be identified was then compared with all other the faces in the previously collected and labelled database to determine if that person has been registered before or not.
- If there was a match, a direct success attendance message will appear on the screen connected to NVIDIA Jetson Nano to display the image from the Raspberry Pi Camera. Otherwise, an accuracy recognition rate message will appear.
- Finally, when the recognition/attendance process is successful, the image of the successful recognition/attendance will be sent to the server to record the recognition/attendance log. The person who needs to be recognized/attend can track the history on the statistics website.

Prior to capturing input frames using a Raspberry Pi Camera, the workflow involves several steps, including face detection, embedded computing, and comparison of the vector to the database using a voting method, as illustrated in Fig. 6. Additionally, a deep neural network was utilized to compute a 128-d vector to quantify each face image in the database.

- The unique ID of the person need to be identified is extracted from the file path.
- The image is loaded and converted into its Gray components.
- Computing of face embedded and adding it to know encodings.

Generating 128 measurements from each face is a crucial step in the facial recognition process, as any image of the same individual should produce consistent measurements. This process requires gathering measurements from each individual's face to create a known-face database. To achieve this, a dictionary containing known-face was created and stored in a data variable before being exported, the encoding was saved in a Yaml file with the 128-d embedding for each face in the dataset. The Haar Cascade algorithm was utilized for face detection, localization, and recognition. The implementation of this algorithm involved two modules from imutils and OpenCV2. Within the video stream from the Raspberry Pi Camera, the image frames were continuously looped over, and the block of code below was executed.

- Load the facial encoding dataset.
- Instantiate the face detector using Haar Cascade.
- Initialise the video stream using the Raspberry Pi Camera connected to the NVIDIA Jetson Nano.
- Wait for the Raspberry Pi Camera to boot up.
- Start capturing frames per second.

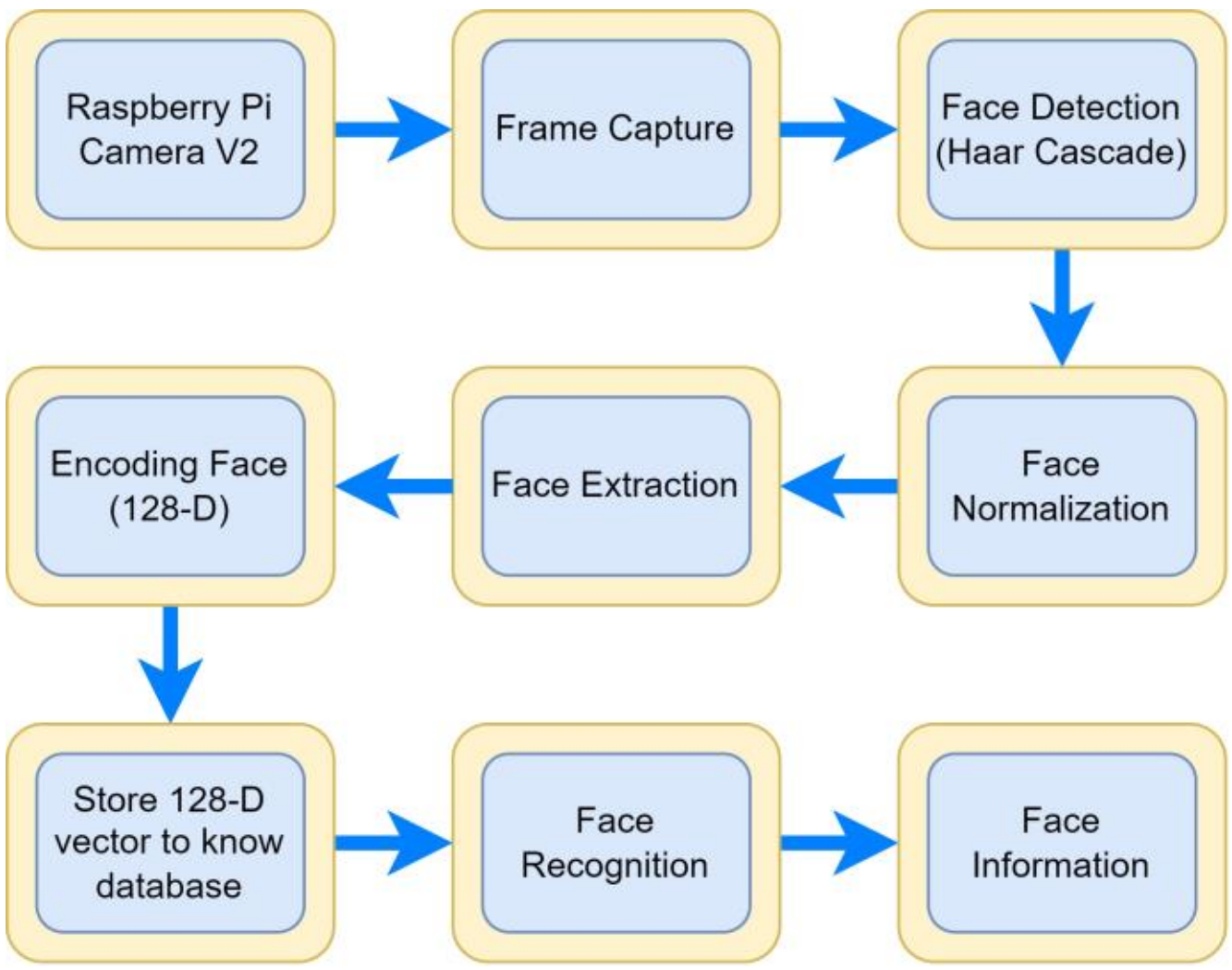

Fig. 6. Face Recognition algorithm utilizing Haar Cascade for Attendance.

*4.3.1. Capturing frames and recognizing faces*

To capture frames from a live video stream, a specific frame is chosen and pre-processed. We utilize the Haar Cascade face detection algorithm, a pre-trained face detection algorithm from OpenCV2 that requires parameter adjustments. The Haar Cascade algorithm is a technique for identifying objects in framed images across multiple real-time scales, called the `detectMultiScale` method. The `detectMultiScale` method is characterized by several parameters that must be configured appropriately.

- The maximum number of faces that can be detected within a frame is specified by MiniNeighbours parameter. The face detector produces a rectangular frame, referred to as the `'rect'`, which encloses the detected face within an image frame. The `rect` is enclosed by four closed lines, which assist in locating and tracking the detected face. From this process, the 128-dimensional encodings of the face image are computed for unique identification purposes.

- The MiniSize parameter sets the minimum size of objects in the frame that the algorithm will consider, ignoring any objects that are smaller than the specified size.
- The Scale Factor parameter is used to control the reduction in image size at each scale during the detection process.

*4.3.2. The detector*

Upon passing the image frame and the `frameCenter` as parameters, the `ObjCenter` method converts the `frame` to `grayscale` to achieve an even background for recognition, considering external factors such as illumination and brightness. Following this step, the `detectMultiScale` method of the Haar Cascade detecting method is employed for face detection.

Once the face was detected, the coordinates of its center were calculated and a rectangle was drawn around it using the `'rect'` function for the purpose of tracking. The servo then moves to follow the movements of the detected face once it is bounded by the detector frame, Fig. 7 shows an example of face detection with Raspberry Pi Camera.

Enabling the Raspberry Pi Camera on the NVIDIA Jetson Nano was accomplished, and a Python script was executed to exhibit and stream the video. A frame in the video stream generated by the Raspberry Pi Camera was captured by the code, which was shown in both RGB color and Gray formats.

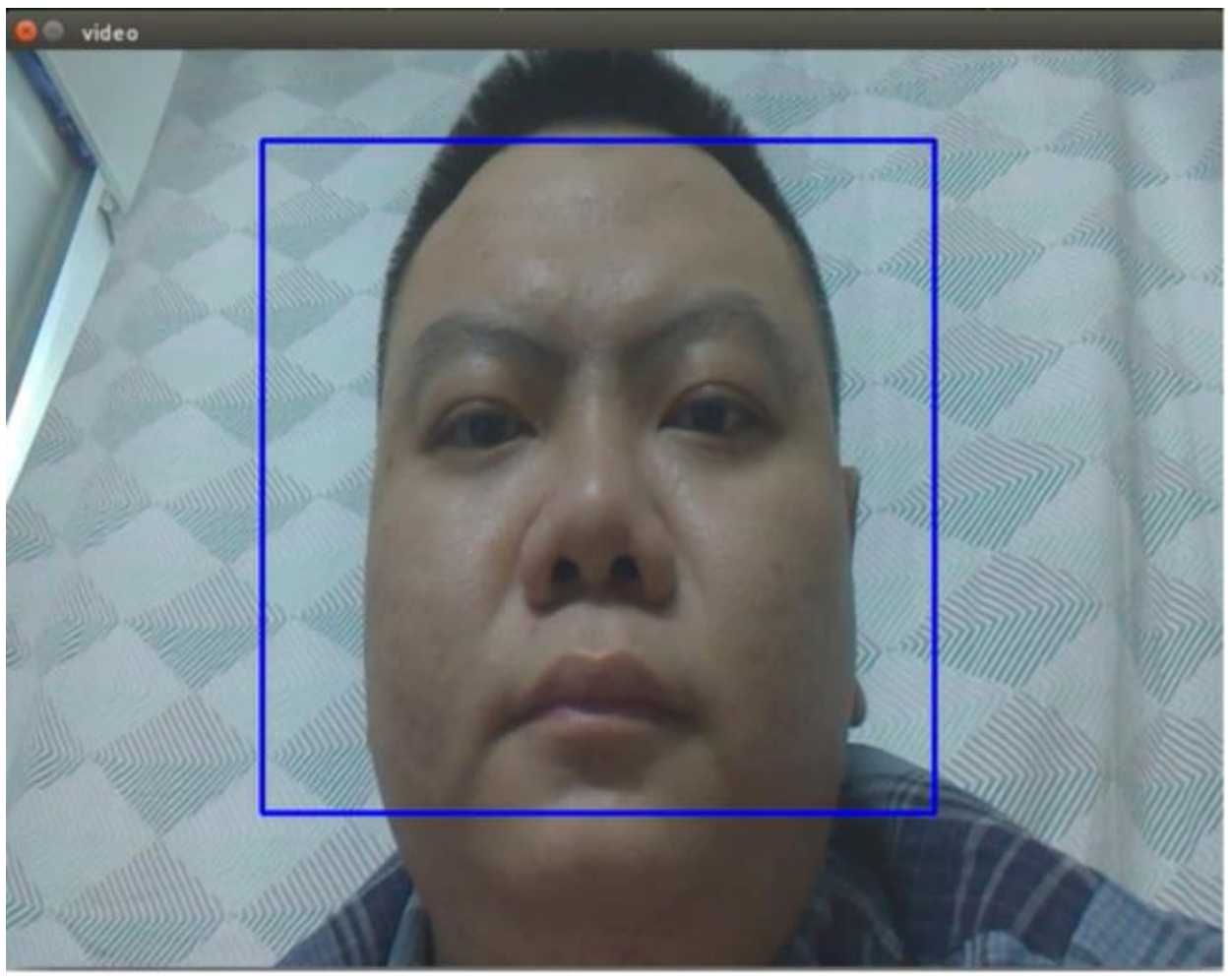

Fig. 7. Face Detection with Raspberry Pi Camera.

*4.3.3. Face detection*

Detecting faces requires the classification of image windows into two categories: those that contain faces (with the surrounding area blurred out) and those that do not (background noise).

- The first step is a classification process that takes an unknown image and outputs a binary value of either yes or no, creating a binary image that shows whether the image contains any faces.
- The second step is the face-positioning task, which takes an image as input and outputs the location of any faces within that image in the form of bounding box with coordinates (`x, y, width, height`). Once a photo has been captured, the system will analyse its database of images for matches and return the most relevant results.

In this study, we will be using the Haar-Like technique to detect faces in real-time video. This technique involves feature selection or extraction for an item in an image using edge detection, line detection, and centered detection to identify specific objects such as eyes, noses, and mouths. This method is used to select the most important features in a picture and extract them for face detection. Once these features have been extracted, we can create a rectangular box in the picture that specifies the location of the face in the image using the x, y, w and h coordinates. We can then save this image as training data for future use. The Haar-Like technique is also used in the recognition process.

A deep learning-based method known as Haar feature-based classifiers was used for object detection in this study. This method involved training the Haar Cascade function on a large number of face images, making it highly accurate due to being trained on both positive and negative images. The function was then applied to detect faces in front of the camera. Algorithm 1 displays the Haar Cascade used for implementing the face detector.

**Algorithm 1:** Face Detection

**Input:** Video Stream
**Output:** Rect. Function bounded face

```
Import libraries
Start Camera ← cv2.VideoCapture
Load cuda_cascadePath ← cv2.cuda_CascadeClassifier
while cam = cv2.VideoCapture do
    Load (ret, img) from camera ← cam.read
    Flip video vertically ← cv2.flip
    Load img_gpu ← cv2.cuda_GpuMat
    Upload img to img_gpu ← cv2.upload
    Convert img_gpu to gray_img_gpu ← cv2.cuda.cvtColor
    Download gray_img from gray_img_gpu ←
    gray_img_gpu.download
    Detect faces ← cuda_cascadePath.detectMultiScale
    for (x, y, w, h) in faces
        Run Rectangle Function ← cv2.rectangle
        Show video ← cv2.imshow
    end for
end while
return Detected Faces
```

*4.3.4. Collecting data*

A folder name dataset was created to store a set of sample faces in grayscale, each with a unique Index. The faces were grouped in thirties, and the dataset had fewer examples, but they were more diverse in terms of pose and lighting. This dataset was used in the next phase to train the Haar Cascade Classifier model for face detection. Algorithm 2 shows the data collecting algorithm used, and a script was written to capture faces and store them in the database.

Algorithm 2 for data acquisition was capable of recording the user’s unique index, which was provided as an integer. Once the face image was captured n number of times, it was stored in dataset directory of the database folder can be seen in Fig. 8. In this case, the algorithm was able to capture three distinct faces, and the outcome of executing the second algorithm.

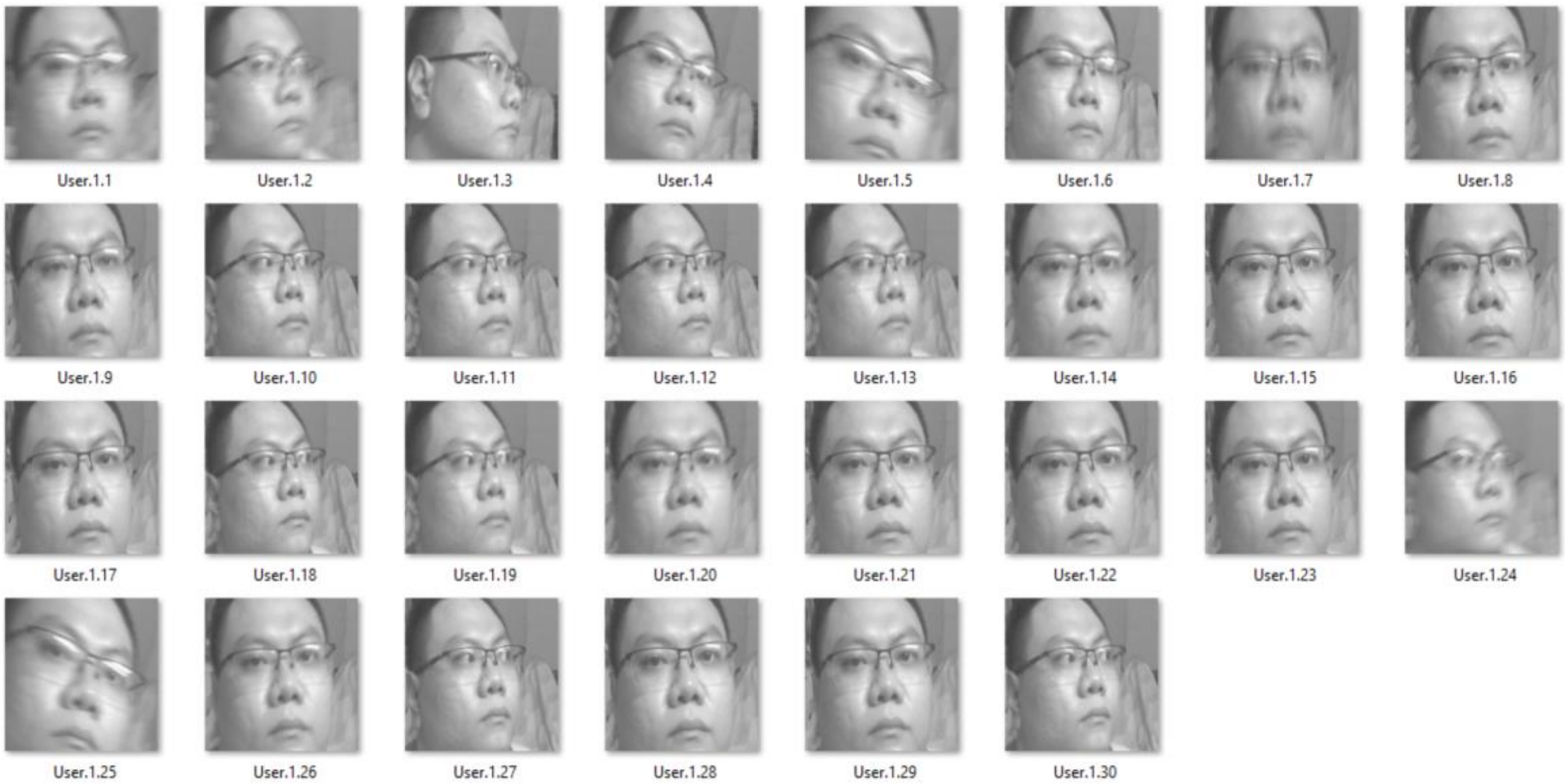

Fig. 8. Dataset of Face-Image for Training.

**Algorithm 2:** Collecting Data

**Input:** Video Stream

**Output:** Gray scaled face image

```
Import libraries
Start Camera ← cv2.VideoCapture
Load cuda_cascadePath ← cv2.cuda_CascadeClassifier
Enter User's Unique Index
Define count = 0
while cam = cv2.VideoCapture do
    Load (ret, img) from camera ← cam.read
    Flip video vertically ← cv2.flip
    Load img_gpu ← cv2.cuda_GpuMat
    Upload img to img_gpu ← cv2.upload
    Convert img_gpu to gray_img_gpu ← cv2.cuda.cvtColor
    Download gray_img from gray_img_gpu ← gray_img_gpu.download
    Detect faces ← cuda_cascadePath.detectMultiScale
    for (x, y, w, h) in faces
        Run Rectangle Function ← cv2.rectangle
        Increase count by 1
        Capture image of face
        Assign [Unique Index].[count+1] to image's name
        Save gray scaled face image to dataset
        Show video ← cv2.imshow
        if count > 30 then
            Break
        end if
    end for
end while
```

*4.3.5. Training the dataset*

To train the face recognition algorithm, we require a dataset comprising of photographs of the people whose identity we intend to recognize. Each image must be associated with a unique index and the order when collecting face samples, allowing the algorithm to identify an input image and output the result accurately. The images of the same person must share the same unique index and be stored in a dataset folder. The folder structure is illustrated in Fig. 9:

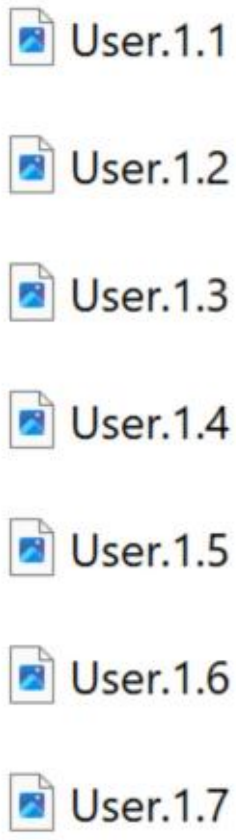

Fig. 9. The Structure of Dataset Folder.

Once the dataset is created, we can proceed with the LBPH (Local Binary Patterns Histograms) computation to generate a training model file. The LBPH method involves creating a binary image that accurately represents the original image, taking into account facial features. This is done using a sliding window approach based on the radius and neighbour parameters.

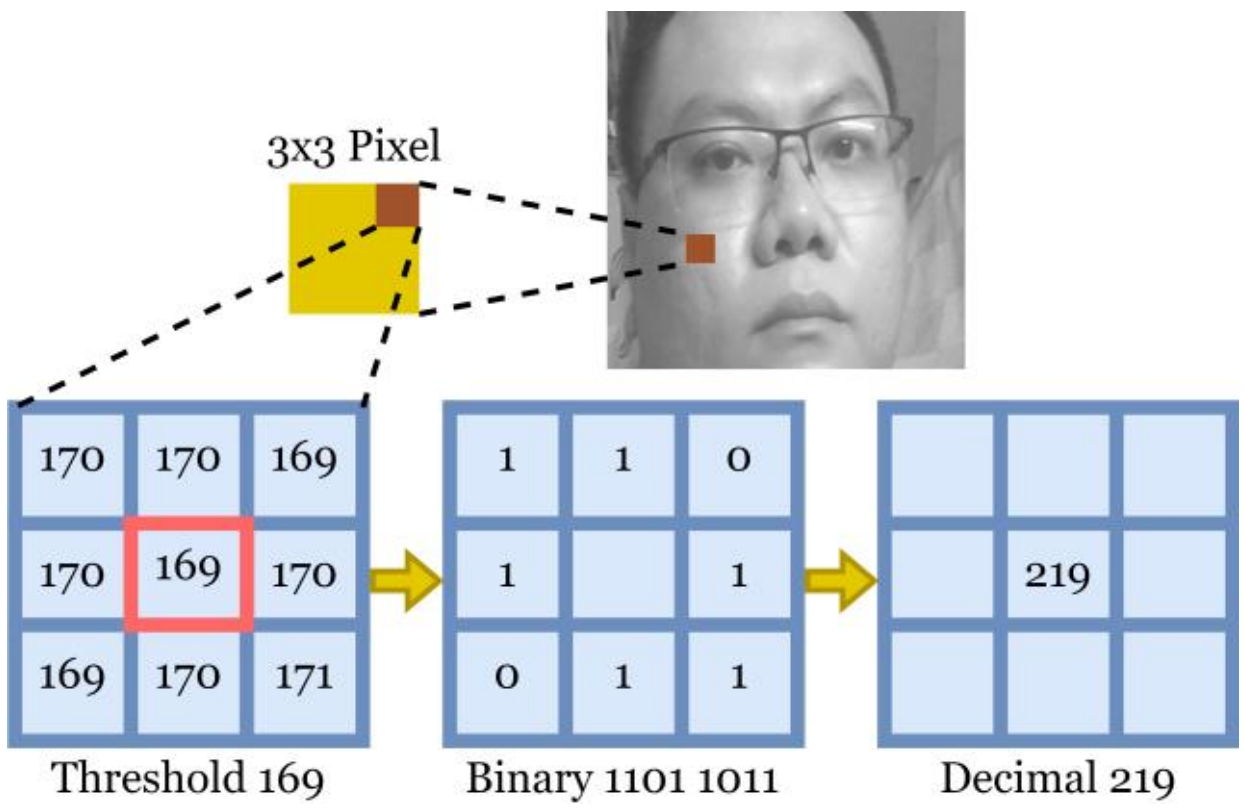

Fig. 10. The LBPH Method.

The `grayscale` image is processed by taking a 3x3 pixel portion of the image as shown in Fig. 10. Each pixel's intensity is represented in a matrix format. The `threshold` value for the center pixel is taken as the reference, and the intensity values of the neighboring pixels are compared with it to calculate the binary values. If the pixel intensity is greater than the threshold value, a binary `"1"` is assigned; otherwise, a `"0"` is assigned. The binary values are combined to form a new binary value, and the final binary value is converted into decimal format. This decimal value replaces the center pixel value. The LBP procedure is applied to all 3x3 grids to get an LBP image, which captures the original image's features more accurately.

We can divide the LBP image into various grids for processing each grid individually. Since a grayscale image only has 256 intensity values (0 to 255), each histogram from each grid will contain 256 places indicating the frequency of each intensity value. These histograms are then combined to form a larger histogram. For instance, if we use a 10x10 grid, the final histogram will have 10x10x256=25600 bins. This histogram effectively captures the original image's properties. Subsequently, we compute and store the histograms for each person into a single file called "`Trainer.yml`" ("`.yml`" represents the file extension for the trained model). Fig. 11 shows "`Trainer.yml`", which is utilized in the recognition process.

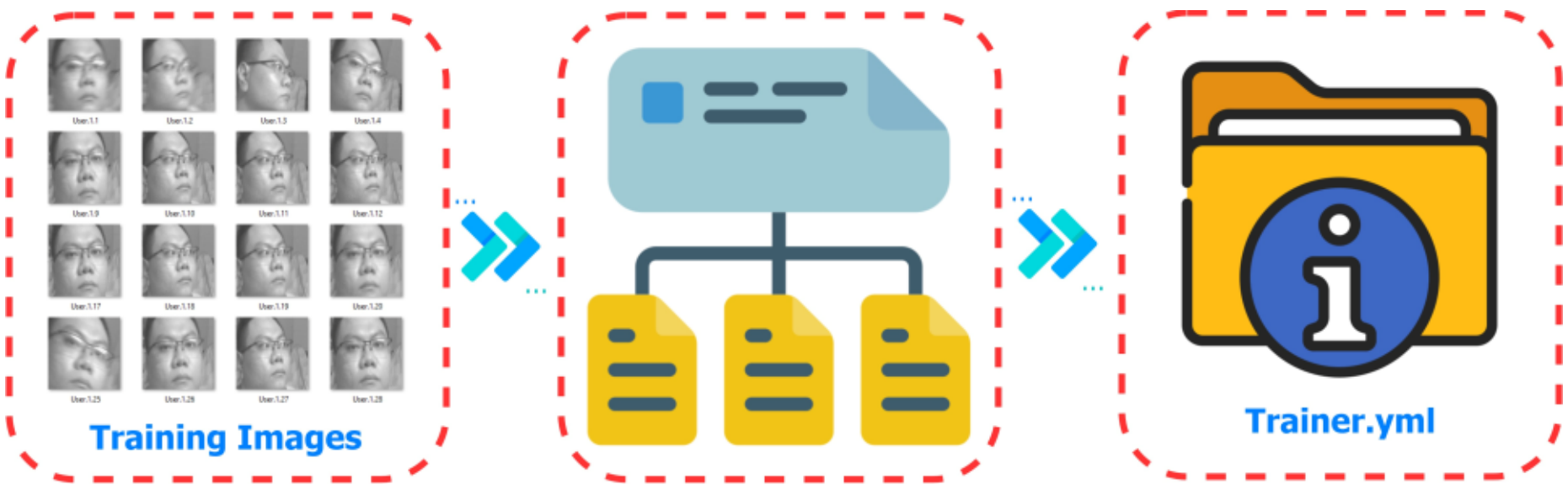

Fig. 11. Trainer.yml is utilized in the recognition process.

### *4.3.6. The trainer*

To teach the OpenCV Recognizer to recognize faces in a real-time video stream, the next step was to train the model by executing Algorithm 3. This algorithm relocated all face images from the dataset folder to the trainer.

**Algorithm 3:** Using Haar Cascade Classifier for Training

**Input:** Gray scaled face images
**Output:** Trained Haar Cascade Classifier

```
Import libraries; Initialisation
Load cuda_cascadePath ← cv2.cuda_CascadeClassifier
Load recognizer ← cv2.LBPHFaceRecognizer_create
Define List of Image Path
Define List of Face Samples
Define List of Unique ID
for Image Path in List of Image Path do
    Load img ← cv2.imread
    Load img_gpu ← cv2.cuda_GpuMat
    Upload img to img_gpu ← cv2.upload
    Convert img_gpu to gray_img_gpu ← cv2.cuda.cvtColor
    Download gray_img from gray_img_gpu ← gray_img_gpu.download
    Detect faces ← cuda_cascadePath.detectMultiScale
    for (x, y, w, h) in faces
        Append Numpy Image to List of Face Samples
        Append Unique ID to List of Unique ID
    end for
    return List of Face Samples; List of Unique ID
end for
Training List of Face Samples and List of Unique
ID with recognizer ← recognizer.train
Write trained results to Trainer.yml
```

The face recognition system employed the LBPH (Local Binary Patterns Histograms) Recognizer, which was accessed through the OpenCV Package. The Rec-ognizer was responsible for processing all images stored in the dataset directory, and it return 2 array of faces and array of unique id. These arrays were then used to train the face recognition algorithm.

*4.3.7. Face recognition*

To recognize a person's face, an external camera is mounted to capture real-time images. The face detection method described earlier is then used to identify faces, and the LBP algorithm is applied to the detected faces to generate a set of histograms. These histograms are compared to the histograms in the previously trained model (`"Trainer.yml"`), and labels for the closest histograms are assigned. Comparison of histograms can be performed using various methods, such as Euclidean distance, absolute value, or chi-square. In this system, the Euclidean distance method is used to determine the closeness of histograms. The label of the image with the closest histogram is returned by the face recognition algorithm, along with the computed distance or "confidence level". Lower confidence levels are preferred because they indicate a closer distance between the two histograms. A threshold value can be set for the confidence level to determine whether the algorithm has correctly recognized the person's face. If the confidence level is below the defined threshold value, the face recognition is considered successful.

*4.4. Privacy and Ethical Considerations*

The current paper focuses on showcasing the practical use of Haar Cascade technology to create an efficient attendance system, particularly in resource-constrained settings. While the paper highlights the system's accuracy and scalability, it has been noted that the aspect of user privacy has not been adequately addressed or explained in the paper. This raises concerns about how the system ensures the privacy of individuals whose attendance data is being collected and processed.

- **Addressing User Privacy:** The preservation of user privacy is of paramount importance in the development and deployment of our innovative attendance system. We understand the ethical and legal obligations associated with handling personal data. To ensure stringent privacy protection, our system adheres to a comprehensive set of data privacy and security measures.
- **Data Anonymization and Encryption:** Central to our privacy strategy is the rigorous application of data anonymization and encryption techniques. Facial data captured during the attendance process undergoes anonymization, rendering it devoid of personally identifiable information. Moreover, robust encryption protocols are employed to guarantee the security of data during both storage and transmission phases.
- **User Consent and Privacy Policy:** The cornerstone of our approach is obtaining explicit user consent. Prior to any facial data collection or storage, individuals are required to provide informed consent. We maintain a transparent privacy policy that comprehensively outlines the data handling procedures. This policy ensures that users are fully aware of how their data will be utilized and protected.
- **Local Processing:** A cornerstone of our approach is local processing on edge devices, notably the NVIDIA Jetson Nano. This design choice minimizes data transmission over potentially vulnerable networks, significantly mitigating privacy concerns.
- **Data Retention Policy:** Our system adheres to a strict data retention policy, which dictates that user data is retained only for the minimum required duration for attendance tracking. Once attendance records are registered, they are promptly anonymized or deleted to uphold user privacy.
- **Consent and Notifications:** In our commitment to transparency, users are presented with clear consent options regarding the use of their facial data for attendance purposes. Additionally, real-time notifications are displayed when facial recognition is in progress, ensuring that users are aware of the system's operations.

User privacy is at the forefront of our ethical considerations. These privacy-enhancing measures are integral to the responsible implementation of our attendance system, underscoring our dedication to safeguarding user data and maintaining the highest ethical standards.

## 5. Experimental Design and Evaluation

The implementation and testing results are presented as follows:

- Accuracy level of the face recognition test.
- Storing attendance history results.
- Analysis of performance.

### *5.1. Result of The Face Recognition Test*

In this section, the recognition script was executed, and the Trainer.yml file was loaded for face recognition. The system continuously monitored in real-time, and all faces that came within the range of the Raspberry Pi Camera were detected and recognized. The recognizer provided a "`prediction`" by returning the person's name label and a confidence level in percentage, indicating how sure the recognizer was that a match had been found.

**Algorithm 4:** Accuracy of Recognition

**Input:** Face Images
**Output:** Accuracy of Recognition

```
Import libraries; Initialisation
Load cuda_cascadePath ← cv2.cuda_CascadeClassifier
Load recognizer ← cv2.LBPHFaceRecognizer_create
Inspect the operation of the camera
while cam = cv2.VideoCapture do
    Flip video vertically ← cv2.flip
    Load img_gpu ← cv2.cuda_GpuMat
    Upload img to img_gpu ← cv2.upload
    Convert img_gpu to gray_img_gpu ← cv2.cuda.cvtColor
    Download gray_img from gray_img_gpu ← gray_img_gpu.download
    Detect faces ← cuda_cascadePath.detectMultiScale
    for (x, y, w, h) in faces
        Run Rectangle Function ← cv2.rectangle
        Run Predict Function ← recognizer.predict
        Get User Unique Index, Confidence
        Convert Confidence value to Percentage
        if Confidence is > 85 then
            Get information of User
        else
            User = Unknown
            Show Confidence
        end if
    end for
end while
return Accuracy of Recognition
```

In this step, instead of using numbered User Unique Indexes, a new array was introduced to display User's Name. This array was trained and matched with the positions of the images. When a face was detected, an important function was called known as the `recognizer.predict()`. This function analysed the captured face and returned the probable owner by indicating the ID name-label and the level of confidence the recognizer had in making the match. If the recognizer was able to predict the face, it would place a text over the image with the probable name label and the percentage probability of the match being correct. In cases where a match could not be made, the recognizer labelled the face as "`unknown`". After running the Recognizer script, the model accurately predicted the names of attendance persons with an average confidence level of 85%. Fig. 12 exhibits the findings of the control tests carried out for the author, demonstrating that the system was able to accurately identify the attendance persons with a precision of 85%. The experimental accuracy of 85% was influenced by various factors such as illumination, facial expression, and the positioning of the face. Our proposed Haar Cascade algorithm (in Algorithm 4) was implemented in OpenCV, a Python-powered platform used for research in Computer Vision, to obtain the simulation results.

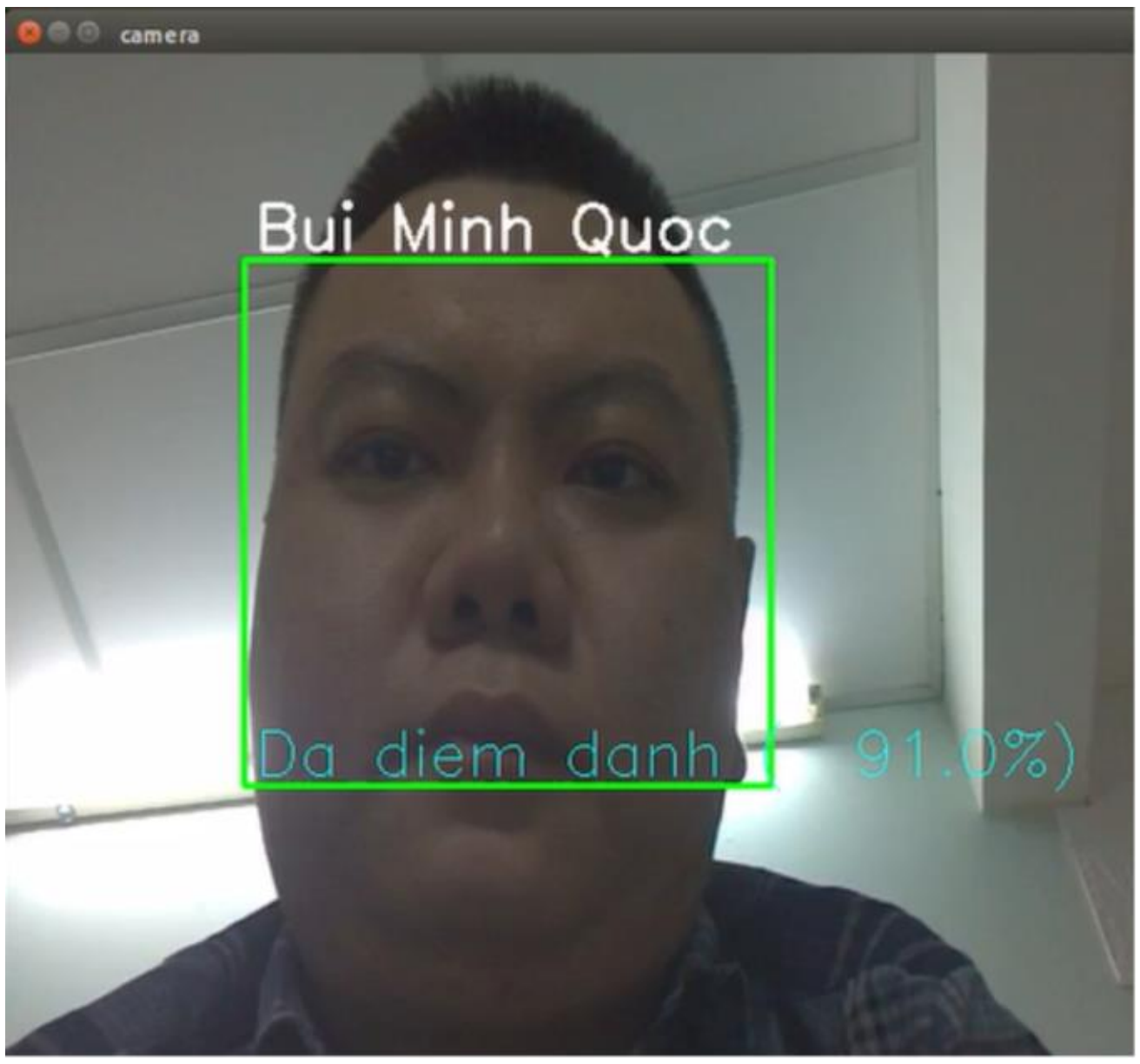

Fig. 12. The accuracy of facial recognition was tested on a control deep-dataset of attendance persons.

### *5.2. The Result of Storing Attendance History*

In this step, the output from the recognizer.py script will be sent to the server with the task of storing the attendance history, including the checked-in images extracted from the Raspberry Pi Camera Module V2 used for recognition. Users can access the system's website to trace the attendance history with information such as date/time of check-in/check-out, images of the checked-in individual, and the location of the NVIDIA Jetson Nano device used for attendance. Additionally, users can search for attendance history within a specific time frame using the criteria of "`From Date – To Date`". The Algorithm 5 displays the algorithm responsible for executing the function that stores attendance history.

**Algorithm 5:** Storing attendance history

**Input:** Gray scaled face images
**Output:** The result of storing attendance history

Import libraries; Initialisation
Run CV2, imwrite
**`while`** Accuracy of Recognition >= Threshold **`do`**
    Store attendance images into the folder containing attendance images ← `cv2.imwrite`
    Convert to gray scale ← `Image.convert`
    Create Multipart Data ← `MultipartEncoder`
    Make Post Request to send Multipart Data to the Server ← `requests.post`
**`end while`**
**`return`** Response Messages from the Server

### 5.3. Performance Analysis

#### 5.3.1. Data gathering

The process involves standing in front of the camera, each member standing in front of a camera and providing their IDs. The process allowed for the input of the number of images to be collected, with two options: 30 images per member or 500 images per member. These datasets were then used to train a face recognition model; this process can be described step-by-step as follows:

- Stand in front of the camera.
- Enter the member's ID.
- Enter the number of images to collect, either 30 or 500 images per member.
- Start the face collection process by following the instructions on the screen.
- When the process is complete, verify the quality of the collected images.
- Repeat steps 2-5 for each team member until all members have their faces collected.

#### 5.3.2. Comprehensive Testing and Training Phases

This section provides a detailed examination of the rigorous testing and training procedures employed in the development of the attendance management system:

- **Dataset Selection:** The dataset used in this evaluation consists of a collection of images containing the faces of individuals who will be subjects of the recognition system. These images serve as the basis for training the facial recognition algorithms to learn the unique features and characteristics of each individual's face.
- **Training Phase:** During the training phase, the selected dataset is used to teach the facial recognition system to recognize and differentiate between different individuals' faces. The system learns to extract relevant facial features and patterns from the images and associates them with corresponding identities.
- **Testing Phase:** Following the training phase, the same dataset is then used for the testing phase. In this phase, the facial recognition system is put to the test by presenting it with a set of images that it has never seen before. These test images contain various scenarios, such as different lighting conditions and the presence of masks, to assess the system's real-world performance.
- **Fairness and Consistency:** By utilizing the same dataset for both training and testing, the evaluation ensures fairness and consistency. This means that both the NVIDIA Jetson Nano and Raspberry Pi 4 are evaluated under identical conditions, facing the same challenges and scenarios. This eliminates potential bias that could arise from using different datasets and allows for a direct, apples-to-apples comparison between the two platforms.

In summary, the use of the same dataset for training and testing serves as a critical element in this evaluation, guaranteeing that the comparison between the NVIDIA Jetson Nano and Raspberry Pi 4 is as fair, consistent, and objective as possible. It ensures that both platforms are assessed based on their ability to recognize faces from the same pool of individuals and under the same set of conditions, providing meaningful insights into their respective performances.

#### 5.3.3. Training

We started team using a dataset of 30 images/member. Each team member collected 30 images of themselves, labeled them with their name or unique index, and combined them into a single dataset folder. The Local Binary Patterns Histograms (LBPH) procedure was applied to all images in the dataset, generating a set of binary images that accurately represented each original image while taking into account facial features. The histograms for each member were computed and stored in a single file called “Trainer.yml” which represented the trained model for the team using a dataset of 30 images/member. The team could then use this file to recognize and classify images of the team members accurately.

For the dataset with 500 images/member, each team member collected 500 images of themselves, labeled them, and combined them into a single dataset folder. The LBPH procedure was applied to all images in the dataset, and the histograms for each member were computed and stored in a Trainer.yml file. The larger dataset size of 500 images/member required more images to process and additional computing resources to train the model effectively.

Both datasets underwent the same Local Binary Patterns Histograms (LBPH) procedure, which was applied to all images in the dataset to generate a set of LBP images. These images were then divided into various grids for individual processing, with each histogram from each grid containing 256 bins that represent the frequency of each intensity value. The resulting histograms are combined to form a larger histogram, effectively capturing the original image's properties. For example, if a 10x10 grid is used, the final histogram will have 10x10x256 = 25600 bins.

In summary, the team using a dataset of 30 images/member and 500 images/member will follow the same training process of collecting images, labeling them, computing LBPH, generating histograms, and storing the histograms in a “Trainer.yml” file. However, the team using a larger dataset size of 500 images/member will have more images to process and may require additional computing resources to train the model effectively.

### *5.3.4. Testing*

We conducted a facial recognition experiment to evaluate the performance of the system in different conditions. We collected facial data for each team member in two datasets sizes, with 30 images and 500 images per member, respectively. The system was trained on these datasets, and we tested its recognition accuracy un-der various conditions.

For the dataset size of 30 images per member, the system achieved an average recognition accuracy of 79% in normal lighting conditions. However, its performance deteriorated in low lighting conditions, with an average recognition accuracy of only 72%. The system also struggled to recognize individuals wearing mask, with an average recognition accuracy of only 65% in normal lighting conditions and 41% in low lighting conditions.

For the dataset size of 500 images per member, the system showed significant improvement in recognition accuracy. It achieved an average recognition accuracy of 93% in normal lighting conditions, but its performance dropped to 88% in low lighting conditions. The system was still able to recognize individuals wearing masks in normal lighting conditions, with an average recognition accuracy of 75%. However, its performance deteriorated further in low lighting conditions, with an average recognition accuracy of only 56%.

Overall, the results indicate that the performance of the facial recognition system was influenced by several factors, including the amount of training data, lighting conditions, and whether or not subjects were wearing masks. The system showed better performance with a larger dataset size and in normal lighting conditions, but its performance deteriorated in low lighting conditions and when subjects were wearing masks.

The testing process provided the following parameters:

- For the case of 30 images/member, the training time was 3,354 seconds under the following conditions:
  - Normal lighting conditions: the average recognition rate is 79% accurate.
  - Low lighting conditions: the average recognition rate is 72% accurate.
  - Wearing a mask and normal lighting conditions: the average recognition rate is 65% accurate.
  - Wearing a mask and low lighting conditions: the average recognition rate is 41% accurate.
- For the case of 500 images/member, the training time was 15,603 seconds under the following conditions:
  - Normal lighting conditions: the average recognition rate is 93% accuracy.
  - Low lighting conditions: the average recognition rate is 88% accuracy.
  - Wearing a mask and normal lighting conditions: the average recognition rate is 75% accuracy.
  - Wearing a mask and low lighting conditions: the average recognition rate is 56% accuracy.

The experimental results are summarized in Table 1.

Table 1. Experimental Results.

| Test case | Training time | Conditions/Average accuracy | Normal Light | Low Light |
|---|---|---|---|---|
| 30 images/member | 3.354 seconds | Normal | 79% | 72% |
| | | Wearing Mask | 65% | 41% |
| 500 images/member | 15.603 seconds | Normal | 93% | 88% |
| | | Wearing Mask | 75% | 56% |

### *5.4. Performance Evaluation*

The system is capable of being trained quickly in just 3,354 seconds when using 30 images per member. However, under normal lighting conditions, the average recognition rate is only 79%, which is relatively low. Under low lighting conditions or when the person is wearing a mask, the system completely fails to recognize the person, which indicates that the system has limitations under these conditions.

When the number of images per member is increased to 500, the training time is longer, taking 15,603 seconds to complete. However, the system's accuracy is generally better, with an average recognition rate of 93% under normal lighting conditions and 88% under low lighting conditions. Even when the person is wearing a mask under normal lighting conditions, the system still performs reasonably well with an average recognition rate of 75%. When the person is wearing a mask and under low lighting conditions, the average recognition rate drops to 56%, which is still significantly better than the previous result of completely failing to recognize the person.

Overall, the system's training performance is limited under certain conditions, such as low lighting conditions or when the person is wearing a mask, but it per-forms relatively well under normal lighting conditions. It's worth noting that when the system starts the camera and executes the recognition task by streaming video from the Raspberry Pi Camera, the CPU usage is over 70% for each core, and it consumes 2.4GB of RAM on the device. Therefore, further improvements or alternative approaches may be necessary to increase the system's accuracy and reliability.

### *5.5. Comparative Analysis*

NVIDIA Jetson Nano and Raspberry Pi 4 are two popular single-board computers that are widely used in a variety of applications, including robotics, AI, and IoT. While both devices offer powerful computing capabilities at an affordable price, there are several key differences between them. The Jetson Nano is designed specifically for AI and machine learning tasks, with a powerful GPU that can handle complex computations. In contrast, the Raspberry Pi 4 has a more general-purpose design, with a CPU that is optimized for everyday computing tasks. Additionally, the Jetson Nano supports more advanced AI and machine learning frameworks out of the box, while the Raspberry Pi 4 requires additional configuration and installation of these frameworks. Ultimately, the choice be-tween the Jetson Nano and Raspberry Pi 4 will depend on the specific requirements of the project, as well as the user's budget and expertise in programming and hardware configuration. The comparative specifications of the two devices are summarized in Table 2.

Table 2. Hardware Specification Comparison

| Specification | NVIDIA Jetson Nano | Raspberry Pi 4 |
|---|---|---|
| CPU | Quad-core ARM Cortex-A57 64-bit @ 1.43 GHz | Quad-core 64-bit ARM Cortex-A72 @ 1.5 GHz |
| GPU | 128-Core NVIDIA Maxwell | Broadcom VideoCore VI |
| RAM | 4GB LPDDR4 | 4GB LPDDR4 |

Recognizing the significance of conveying complex performance metrics in an accessible manner, we have introduced a comprehensive and visually rich performance comparison section. Central to this section is Table 3, which provides an at-a-glance summary of key performance indicators for the NVIDIA Jetson Nano and Raspberry Pi 4. This table provides a comprehensive comparison of the performance metrics between two embedded computing platforms, namely the NVIDIA Jetson Nano and Raspberry Pi 4, when employed for the implementation of a facial recognition system. Both platforms utilize the same algorithms and libraries, including Haar Cascade for face detection, Python as the programming language, OpenCV2 for image processing, LBPHFaceRecognizer_create for face recognition, and cuda_CascadeClassifier for GPU acceleration.

The experimental process on Raspberry Pi 4 using the same algorithm as on NVIDIA Jetson Nano, and obtain the following experimental results:

- **Recognition Accuracy (%):** The recognition accuracy is a crucial indicator of a facial recognition system's effectiveness. In the 30 Images/Member scenario, both the NVIDIA Jetson Nano and Raspberry Pi 4 exhibited similar recognition accuracy under normal lighting conditions, achieving a rate of 78% and 65%, respectively. However, the recognition accuracy dropped significantly in challenging conditions. When subjected to low lighting conditions, both systems faced decreased accuracy, with the Jetson Nano at 65% and the Raspberry Pi 4 at 30%. Furthermore, both systems failed to recognize individuals wearing masks. Upon increasing the number of training images per member to 500, both platforms showed improvement in recognition accuracy. The NVIDIA Jetson Nano achieved an average recognition rate of 93% under normal lighting conditions and 88% under low lighting conditions. Even in the presence of masks under normal lighting conditions, it maintained a commendable accuracy rate of 75%. In low lighting conditions with masks, the accuracy dropped to 56%. Meanwhile, the Raspberry Pi 4 achieved an accuracy of 76% under normal lighting conditions, which was slightly lower than the Jetson Nano. In low lighting conditions, the accuracy declined to 52%, and masks caused recognition failure, similar to the Jetson Nano.
- **Training Time (seconds):** The training time required for each platform provides insights into their computational efficiency. In the 30 Images/Member scenario, the NVIDIA Jetson Nano completed the training phase in 3,354 seconds, while the Raspberry Pi 4 required approximately twice as long, with a training time of 16,695 seconds. This significant difference in training time highlights the computational advantages of the Jetson Nano. With an increased dataset of 500 Images/Member, the Jetson Nano took 15,603 seconds to complete the training, whereas the Raspberry Pi 4 extended its training time to 43,201 seconds. These results underline the substantial impact of dataset size on training time and further emphasize the Jetson Nano's computational efficiency.
- **Resource Utilization:** Both platforms exhibited high CPU usage during the execution of the recognition task. Each core of both the Jetson Nano and Raspberry Pi 4 operated at over 70% utilization, indicating the computational intensity of the face recognition process. Additionally, both platforms consumed 2.4GB of RAM, highlighting the memory requirements of the system. This resource-intensive operation is attributed to the complex calculations involved in image processing and recognition.
- **Frames per Second (FPS):** Frames per second (FPS) measure the real-time processing capabilities of each platform. In the 30 Images/Member scenario, the NVIDIA Jetson Nano achieved a frame rate of 20 FPS, indicating its ability to process video streams and perform facial recognition tasks efficiently. Conversely, the Raspberry Pi 4 operated at a lower frame rate of 10 FPS, suggesting slightly reduced real-time processing capabilities compared to the Jetson Nano. In the 500 Images/Member scenario, the NVIDIA Jetson Nano demonstrated further improved performance, achieving a frame rate of 30 FPS, while the Raspberry Pi 4 continued to operate at a lower frame rate of 15 FPS.

The summary of this comprehensive performance comparison underscores the recognition accuracy, training time, resource utilization, and real-time processing capabilities of the NVIDIA Jetson Nano and Raspberry Pi 4. It also emphasizes the significance of dataset size in training time and underscores the superior computational efficiency and real-time performance of the NVIDIA Jetson Nano in face recognition tasks, making it an ideal choice for applications demanding accuracy and speed as shown in Table 3.

Table 3. The summary of the comparison results

| **Test Case** | **NVIDIA Jetson Nano** | **Raspberry Pi 4** |
|---|---|---|
| **30 images/member training time** | **3.354 seconds** | **16.695 seconds** |
| The average of accuracy of Normal lighting conditions | 78% | 65% |
| The average of accuracy of Normal lighting conditions and Wearing mask | 65% | - |
| The average of accuracy of Low lighting conditions | 72% | 30% |
| The average of accuracy of Low lighting conditions and Wearing mask | 41% | - |
| Resource utilization | GPU-accelerated | CPU-based |
| Frames per Second (FPS) | 20 FPS | 10 FPS |
| **500 images/member training time** | **15.603 seconds** | **43.201 seconds** |
| The average of accuracy of Normal lighting conditions | 93% | 76% |
| The average of accuracy of Normal lighting conditions and Wearing mask | 75% | 43% |
| The average of accuracy of Low lighting conditions | 88% | 52% |
| The average of accuracy of Low lighting conditions and Wearing mask | 56% | - |
| Resource utilization | GPU-accelerated | CPU-based |
| Frames per Second (FPS) | 30 FPS | 15 FPS |

Based on the above results, it can be concluded that using the NVIDIA Jetson Nano will provide significantly better performance than using the Raspberry Pi 4 for facial recognition tasks. The Jetson Nano was able to achieve higher accuracy rates in both normal and low lighting conditions, and with or without the presence of a mask, compared to the Raspberry Pi 4. Additionally, the training time for the Jetson Nano was shorter than that of the Raspberry Pi 4.

Moreover, the training time on Jetson Nano is much shorter than that on Raspberry Pi 4 for both cases, which indicates that Jetson Nano is more efficient in training the face recognition model.

The NVIDIA Jetson Nano performs significantly better than the Raspberry Pi for this application, producing results that are nearly twice as good. One of the main reasons for this is that the Jetson Nano supports GPU acceleration through the use of CUDA. The CUDA technology enables the Jetson Nano to accelerate complex calculations, which leads to faster and more efficient processing of data. This allows the system to offload many of the compute-intensive tasks involved in image processing and recognition to the GPU, freeing up the CPU to handle other tasks. As a result, the Jetson Nano is able to process video streams and perform facial recognition tasks much more quickly and accurately than the Raspberry Pi. Additionally, the Jetson Nano has more powerful hardware specifications overall, including a higher-end CPU, more RAM, and more storage, all of which contribute to its superior performance. Overall, the Jetson Nano is a great choice for developers looking to build high-performance computer vision applications. On the other hand, the Raspberry Pi only relies on its CPU for processing, which can lead to longer processing times and lower accuracy in comparison to the NVIDIA Jetson Nano. This is because the CPU is not specifically designed for complex parallel computations like GPUs are. Therefore, the NVIDIA Jetson Nano's ability to utilize a GPU with CUDA significantly enhances its processing power and makes it a much better choice for applications that require real-time video processing and high accuracy recognition.

Therefore, based on the experimental results, it can be concluded that the recognition accuracy of the system can be significantly improved by increasing the number of images per member during the training phase. Additionally, the system performs better under normal lighting conditions compared to low lighting conditions, and wearing a mask can also negatively impact its accuracy. For applications that require high performance and accuracy, the NVIDIA Jetson Nano is a better choice over the Raspberry Pi 4 due to its GPU support with CUDA. However, the decision to choose between these two hardware options also depends on other factors such as cost, availability, and specific project requirements. Therefore, it is essential to carefully consider all of these factors when deciding on the appropriate hardware for a facial recognition system.

## 6. Discussion

### *6.1. Contributions to Literature*

The integration of face recognition technology in attendance management, as exemplified by our Face Recognition Attendance Management System (FRAMS), represents a significant leap forward in this domain. The system, which synergizes the NVIDIA Jetson Nano, Raspberry Pi Camera, and the Haar Cascade classifier, demonstrates a remarkable fusion of cutting-edge hardware and software technologies.

Our research contributes to the literature in several ground-breaking ways:

- **Technological Synergy:** By integrating the NVIDIA Jetson Nano with the Raspberry Pi Camera and Haar Cascade classifiers, FRAMS pushes the boundaries of what is technically feasible in biometric attendance systems. Previous studies (Y. Wen et al., 2021; Mani et al., 2021; Singh et al., 2021; Salih et al., 2020; Adoghe et al., 2021; Chandramouli, 2021; Nayak et al., 2021; Mittal et al., 2022; Votto et al., 2021; Khan et al., 2021; Herath et al., 2022; Lata et al., 2022; Ray et al., 2023; Jamwal et al., 2022; Pathare et al., 2023; Dr. Varsha 2023) have separately explored the capabilities of these technologies, but our work is among the first to merge them in an attendance management context, showcasing enhanced processing speed and reliability even under varied environmental conditions.
- **Advancement in Real-World Application:** Our system addresses critical gaps identified in existing research regarding the robustness and adaptability of facial recognition technologies. By successfully implementing these technologies in a real-world attendance system, we provide empirical evidence supporting their viability and efficiency, thereby expanding the theoretical framework for future biometric studies.
- **Methodological Innovation:** The use of Haar Cascade algorithms in conjunction with advanced computing hardware like the NVIDIA Jetson Nano represents a significant methodological advancement. This combination allows for the processing of complex video data streams with high accuracy and minimal latency, a critical improvement over existing models that often struggle with speed and accuracy in variable lighting conditions.
- **Benchmarking New Standards:** By documenting the specific performance metrics of FRAMS, such as recognition accuracy, processing speed, and system scalability, our study sets new benchmarks for future research. These metrics provide a tangible framework that other researchers can aim to meet or exceed, fostering a competitive and innovative research environment.
- **Interdisciplinary Contributions:** Introducing a new interdisciplinary area of study that combines technical efficiency with organizational behavior. This integration not only broadens the applicative scope of facial recognition technologies but also encourages collaboration between technologists and business management professionals.

By grounding our technological innovations within the context of existing academic discussions and highlighting their implications for both theory and practice, this research substantially enriches the academic literature on biometric technologies and their applications in diverse organizational settings.

### *6.2. Practical Implications*

The practical implications of FRAMS are vast, offering enhanced operational efficiencies in various sectors. In educational settings, the system reduces administrative burdens by automating attendance tracking, thereby allowing institutions to better allocate resources and focus on educational quality. For workplaces, FRAMS offers a method to streamline workforce management, enhancing productivity by minimizing the time required for attendance processing. Additionally, the system's scalability makes it an ideal solution for managing large-scale events, ensuring accurate and efficient guest management while also enhancing security measures. These applications demonstrate the system's potential to transform attendance management practices by integrating facial recognition technology effectively.

At the core of our research is a commitment to revolutionize attendance tracking by harnessing the potential of facial recognition technology. Our findings reveal several key aspects:

- **Efficiency and Robustness:** FRAMS exhibits impressive efficiency in processing video streams, effectively handling variable lighting conditions and pose variations. This efficiency is underpinned by the utilization of Haar

Cascade and OpenCV2, which contribute to the system's low computational requirements, ensuring its practical applicability in real-world scenarios.

- **Innovative Integration:** The incorporation of the NVIDIA Jetson Nano signifies a breakthrough in merging advanced technology with affordability and energy efficiency. This integration addresses a gap in existing literature, where such a combination is seldom explored.
- **Distinctive Advantages:** FRAMS excels in various facets:
  - **Accuracy:** Ensuring precise attendance records with minimal errors.
  - **Speed:** Offering rapid processing for efficient attendance tracking.
  - **Scalability and Adaptability:** Suitable for different scales and contexts, from educational settings to corporate environments and event management.
  - **Cost-Efficiency:** Identifying avenues for cost optimization to make our system even more accessible to a wider array of institutions.

FRAMS not only advances the technical landscape of attendance systems but also brings substantial practical benefits to various sectors. These implications can be articulated as follows:

- **Health and Safety:** In light of global health concerns such as pandemics, FRAMS can play a pivotal role in non-contact attendance taking, minimizing physical interactions and potentially reducing the spread of infections in large gatherings.
- **Environmental Impact:** By reducing the need for physical documentation and traditional attendance methods, FRAMS contributes to environmental sustainability efforts, reducing paper use and waste.
- **Regulatory Compliance:** For sectors with stringent regulatory requirements for attendance and personnel tracking, such as healthcare and finance, FRAMS offers a compliant, reliable, and auditable system that can adapt to various regulatory environments.
- **Technological Democratization:** The cost-effective nature of the NVIDIA Jetson Nano enables smaller institutions and those in developing regions to adopt cutting-edge technology, democratizing access to advanced technological solutions and leveling the playing field in technological adoption.
- **Social and Ethical Considerations:** As facial recognition technology becomes more pervasive, it is imperative to address the social and ethical implications of its deployment. Our system is designed with built-in privacy safeguards to protect individuals' data and adhere to GDPR standards. The ethical deployment of FRAMS also involves transparent data usage policies and regular audits to ensure compliance with international privacy laws. By proactively addressing these concerns, FRAMS not only enhances security and efficiency but also respects the privacy and rights of all individuals, paving the way for responsible use of technology in sensitive environments.

The deployment of FRAMS has profound implications across various sectors. Educational institutions can benefit from streamlined attendance processes, while workplaces can use the system for enhanced workforce management. Additionally, event organizers can utilize FRAMS for guest management and heightened security. By addressing these practical aspects, FRAMS not only meets the immediate needs of efficient and accurate attendance management but also sets a precedent for future technological advancements in various administrative functions.

## 7. Conclusion and Future Research

Our Face Recognition Attendance Management System (FRAMS) stands as a pioneering solution in the field of attendance management. It uniquely blends technology, efficiency, and cost-effectiveness, addressing the existing gaps in attendance tracking systems. The system's high accuracy, speed, and scalability make it a versatile tool across multiple sectors, promising significant operational efficiencies and potential cost savings.

As we look to the future, the potential for FRAMS to evolve is vast. Anticipated enhancements include:

- **Enhanced Security:** Strengthening the system against emerging vulnerabilities.
- **Feature Enrichment:** Adding functionalities such as emotion recognition or mask detection to increase utility.
- **IoT Integration:** For seamless data sharing and analytics.
- **User-Centric Design:** Ensuring the system caters to diverse user groups, including those with disabilities.

In conclusion, FRAMS is more than just an attendance system; it represents the vast potential of technological innovation in solving real-world problems. Its ongoing development will continue to contribute to the advancement of attendance management systems. Our work highlights the significance of interdisciplinary research and development in creating solutions that are not only technologically advanced but also socially relevant and practically viable.

### Declaration of interests

The authors declare that they have no known competing financial interests or personal relationships that could have appeared to influence the work reported in this paper.

### Acknowledgement

This research was supported by The VNUHCM-University of Information Technology's Scientific Research Support Fund.